# Conditional Plausibility Measures and Bayesian Networks

**Joseph Y. Halpern**                                    HALPERN@CS.CORNELL.EDU
*Cornell University, Computer Science Department*
*Ithaca, NY 14853*
*http://www.cs.cornell.edu/home/halpern*

## Abstract

A general notion of algebraic conditional plausibility measures is defined. Probability measures, ranking functions, possibility measures, and (under the appropriate definitions) sets of probability measures can all be viewed as defining algebraic conditional plausibility measures. It is shown that algebraic conditional plausibility measures can be represented using Bayesian networks.

## 1. Introduction

Pearl (1988) among others has long argued that Bayesian networks (that is, the dags without the conditional probability tables) represent important *qualitative* information about uncertainty regarding conditional dependencies and independencies. To the extent that this is true, Bayesian networks should make perfect sense for non-probabilistic representations of uncertainty. And, indeed, Bayesian networks have been used with $\kappa$ rankings (Spohn, 1988) by Darwiche and Goldszmidt (1994). It follows from results of Wilson (1994) that *possibility measures* (Dubois & Prade, 1990) can be represented using Bayesian networks.

The question I address in this paper is "What properties of a representation of uncertainty are required to be able to represent the uncertainty using a Bayesian network?" This question too has been addressed in earlier work, see (Darwiche, 1992; Darwiche & Ginsberg, 1992; Friedman & Halpern, 1995; Wilson, 1994), although the characterization given here is somewhat different. Shenoy and Shafer (1990) consider a related question—essentially, what is required of a representation of uncertainty so that marginals can be computed using "local computations" of the type used in Bayesian networks—and provide axioms sufficient to guarantee that this is possible.

Here I represent uncertainty using *plausibility measures*, as in (Friedman & Halpern, 1995). To answer the question, I must examine general properties of conditional plausibility as well as defining a notion of plausibilistic independence. Unlike earlier papers, I enforce a symmetry condition in the definition of conditional independence, so that, for example, $A$ is independent of $B$ iff $B$ is independent of $A$. While this property holds for probability, under the asymmetric definition of independence used in earlier work it does not necessarily hold for other formalisms. There are also subtle but important differences between this paper and (Friedman & Halpern, 1995) in the notion of conditional plausibility. The definitions here are simpler but more general; particular attention is paid here to conditions on when the conditional plausibility must be defined.

The major results here are a general condition, simpler than that given in (Friedman & Halpern, 1995; Wilson, 1994), under which a conditional plausibility measure satisfies the semi-graphoid properties (which means it can be represented using a Bayesian network). Conditions are also given that suffice for a Bayesian network to be able to quantitatively represent a plausibility measure; more precisely, conditions are given so that a plausibility measure can be uniquely reconstructed given conditional plausibility tables for each node in the Bayesian network. Conditions for quantitative representation by Bayesian networks do not seem to have been presented in the literature for representations of uncertainty other than probability (for which the conditions are trivial). A minor additional condition also suffices to guarantee that *d-separation* in the network characterizes conditional independence. All these conditions clearly apply to $\kappa$ rankings and possibility measures. Perhaps more interestingly, they also apply to sets of probabilities under a novel representation of such sets as a plausibility measure. This novel representation (and the associated notion of conditioning) is shown to have some natural properties not shared by other representations.

The rest of the paper is organized as follows. In Section 2, I discuss conditional plausibility measures. Section 3 introduces *algebraic* conditional plausibility measures, which are ones where there is essentially an analogue to + and ×. (Putting such an algebraic structure on uncertainty is not new; it was also done in (Darwiche, 1992; Darwiche & Ginsberg, 1992; Friedman & Halpern, 1995; Weydert, 1994).) Section 4 discusses independence and conditional independence in conditional plausibility spaces, and shows that algebraic conditional plausibility measures satisfy the semi-graphoid properties. Finally, in Section 5, Bayesian networks based on (algebraic) plausibility measures are considered. Combining the fact that algebraic plausibility measures satisfy the semi-graphoid properties with the results of (Geiger, Verma, & Pearl, 1990), it follows that d-separation in a Bayesian network $G$ implies conditional independence for all algebraic plausibility measures compatible with $G$; a weak richness condition is shown to yield the converse. The paper concludes in Section 6. Longer proofs are relegated to the appendix.

## 2. Conditional Plausibility

### 2.1 Unconditional Plausibility Measures

Before getting to conditional plausibility measures, it is perhaps best to consider unconditional plausibility measures. The basic idea behind plausibility measures is straightforward. A probability measure maps subsets of a set $W$ to $[0, 1]$. Its domain may not consist of all subsets of $W$; however, it is required to be an *algebra*. (Recall that an algebra $\mathcal{F}$ over $W$ is a set of subsets of $W$ containing $W$ and closed under union and complementation, so that if $U, V \in \mathcal{F}$, then so are $U \cup V$ and $\overline{U}$.) A *plausibility measure* is more general; it maps elements in an algebra $\mathcal{F}$ to some arbitrary partially ordered set. If Pl is a plausibility measure, then we read $\mathrm{Pl}(U)$ as "the plausibility of set $U$". If $\mathrm{Pl}(U) \leq \mathrm{Pl}(V)$, then $V$ is at least as plausible as $U$. Because the ordering is partial, it could be that the plausibility of two different sets is incomparable. An agent may not be prepared to say of two sets that one is more likely than another or that they are equal in likelihood.

Formally, a *plausibility space* is a tuple $S = (W, \mathcal{F}, \mathrm{Pl})$, where $W$ is a set of worlds, $\mathcal{F}$ is an algebra over $W$, and Pl maps sets in $\mathcal{F}$ to some set $D$ of *plausibility values* partially ordered by a relation $\leq_D$ (so that $\leq_D$ is reflexive, transitive, and anti-symmetric) that contains two special elements $\top_D$ and $\bot_D$ such that $\bot_D \leq_D d \leq_D \top_D$ for all $d \in D$; these are intended to be the analogues of 1 and 0 for probability. As usual, the ordering is defined $<_D$ by taking $d_1 <_D d_2$ if $d_1 \leq_D d_2$ and $d_1 \neq d_2$. I omit the subscript $D$ from $\leq_D$, $<_D$, $\top_D$ and $\bot_D$ whenever it is clear from context.

There are three requirements on plausibility measures. The first two are obvious analogues of requirements that hold for other notions of uncertainty: the whole space gets the maximum plausibility and the empty set gets the minimum plausibility. The third requirement says that a set must be at least as plausible as any of its subsets.

Pl1. $\text{Pl}(\emptyset) = \bot_D$.

Pl2. $\text{Pl}(W) = \top_D$.

Pl3. If $U \subseteq U'$, then $\text{Pl}(U) \leq \text{Pl}(U')$.

(In Pl3, I am implicitly assuming that $U, U' \in \mathcal{F}$. Similar assumptions are made throughout.)

All the standard representations of uncertainty in the literature can be represented as plausibility measures. I briefly describe some other representations of uncertainty that will be of relevance to this paper.

**Sets of probabilities:** One common way of representing uncertainty is by a set of probability measures. This set is often assumed to be convex (see, for example, (Campos & Moral, 1995; Cousa, Moral, & Walley, 1999; Gilboa & Schmeidler, 1993; Levi, 1985; Walley, 1991) for discussion and further references), however, convex sets do not seem appropriate for representing independence assumptions, so I do not make this restriction here. For example, if a coin with an unknown probability of heads is tossed twice, and the tosses are known to be independent, it seems that a reasonable representation is given by the set $\mathcal{P}_0$ consisting of all measures $\mu_\alpha$, where $\mu_\alpha(hh) = \alpha^2$, $\mu_\alpha(ht) = \mu_\alpha(th) = \alpha(1-\alpha), \mu_\alpha(tt) = (1-\alpha)^2$. Unfortunately, $\mathcal{P}_0$ is not convex. Moreover, its convex hull includes many measures for which the coin tosses are not independent. It is argued in (Cousa et al., 1999) that a set of probability measures is behaviorally equivalent to its convex hull. However, even if we accept this argument, it does not follow that a set and its convex hull are equivalent insofar as determination of independencies goes.

There are a number of ways of viewing a set $\mathcal{P}$ of probability measures as a plausibility measure. One uses the *lower probability* $\mathcal{P}_*$, defined as $\mathcal{P}_*(U) = \inf\{\mu(U) : \mu \in \mathcal{P}\}$. Clearly $\mathcal{P}_*$ satisfies Pl1–3. The corresponding *upper probability* $P^*$, defined as $\mathcal{P}^*(U) = \sup\{\mu : \mu \in \mathcal{P}\} = 1 - \mathcal{P}_*(\overline{U})$, is also clearly a plausibility measure.

Both $\mathcal{P}_*$ and $\mathcal{P}^*$ give a way of comparing the likelihood of two subsets $U$ and $V$ of $W$. These two ways are incomparable; it is easy to find a set $\mathcal{P}$ of probability measures on $W$ and subsets $U$ and $V$ of $W$ such that $\mathcal{P}_*(U) < \mathcal{P}_*(V)$ and $\mathcal{P}^*(U) > \mathcal{P}^*(V)$. Rather than choosing between $\mathcal{P}_*$ and $\mathcal{P}^*$, we can associate a different plausibility measure with $\mathcal{P}$ that captures both. Let $D_{\mathcal{P}_*, \mathcal{P}^*} = \{(a, b) : 0 \leq a \leq b \leq 1\}$ and define $(a, b) \leq (a', b')$ iff $b \leq a'$. This puts a partial order on $D_{\mathcal{P}_*, \mathcal{P}^*}$; clearly $\bot_{D_{\mathcal{P}_*, \mathcal{P}^*}} = (0, 0)$ and $\top_{D_{\mathcal{P}_*, \mathcal{P}^*}} = (1, 1)$. Define $\text{Pl}_{\mathcal{P}_*, \mathcal{P}^*}(U) = (\mathcal{P}_*(U), \mathcal{P}^*(U))$. Thus, $\text{Pl}_{\mathcal{P}_*, \mathcal{P}^*}$ associates with a set $U$ two numbers that can be thought of as defining an interval in terms of the lower and upper probability of $U$. It is easy to check that $\text{Pl}_{\mathcal{P}_*, \mathcal{P}^*}(U) \leq \text{Pl}_{\mathcal{P}_*, \mathcal{P}^*}(V)$ if the upper probability of $U$ is less than or equal to the lower probability of $V$. $\text{Pl}_{\mathcal{P}_*, \mathcal{P}^*}$ satisfies Pl1–3, so it is indeed a plausibility measure, but one which puts only a partial order on events.

The trouble with $\mathcal{P}_*$, $\mathcal{P}^*$, and even $\mathrm{Pl}_{\mathcal{P}_*, \mathcal{P}^*}$ is that they lose information. For example, it is not hard to find a set $\mathcal{P}$ of probability measures and subsets $U, V$ of $W$ such that $\mu(U) \leq \mu(V)$ for all $\mu \in \mathcal{P}$ and $\mu(U) < \mu(V)$ for some $\mu \in \mathcal{P}$, but $\mathcal{P}_*(U) = \mathcal{P}_*(V)$ and $\mathcal{P}^*(U) = \mathcal{P}^*(V)$. Indeed, there exists an infinite set $\mathcal{P}$ of probability measures such that $\mu(U) < \mu(V)$ for all $\mu \in \mathcal{P}$ but $\mathcal{P}_*(U) = \mathcal{P}_*(V)$ and $\mathcal{P}^*(U) = \mathcal{P}^*(V)$. If all the probability measures in $\mathcal{P}$ agree that $U$ is less likely than $V$, it seems reasonable to conclude that $U$ is less likely than $V$. However, none of $\mathcal{P}_*$, $\mathcal{P}^*$, or $\mathrm{Pl}_{\mathcal{P}_*, \mathcal{P}^*}$ will necessarily draw this conclusion.

Fortunately, it is not hard to associate yet another plausibility measure with $\mathcal{P}$ that does not lose this important information. For technical convenience that will become clear later, assume that there is some index set $I$ such that $\mathcal{P} = \{\mu_i : i \in I\}$. Thus, for example, if $\mathcal{P} = \{\mu_1, \ldots, \mu_n\}$, then $I = \{1, \ldots, n\}$. Let $D_I = [0, 1]^I$, that is, the functions from $I$ to $[0, 1]$, with the pointwise ordering, so that $f \leq g$ iff $f(i) \leq g(i)$ for all $i \in I$.[1] It is easy to check that $\perp_{D_I}$ is the function $f : I \to [0, 1]$ such that $f(i) = 0$ for all $i \in I$ and $\top_{D_I}$ is the function $g$ such that $g(i) = 1$ for all $i \in I$. For $U \subseteq W$, let $f_U$ be the function such that $f_U(i) = \mu_i(U)$ for all $i \in I$. For example, for the set $\mathcal{P}_0$ of measures representing the two coin tosses (which is indexed by $I\!\!R$), the set $W$ can be taken to be $\{hh, ht, tt, th\}$. Then, for example, $f_{\{hh\}}(\alpha) = \mu_\alpha(hh) = \alpha^2$ and $f_{\{ht, tt\}}(\alpha) = 1 - \alpha$.

It is easy to see that $f_\emptyset = \perp_{D_I}$ and $f_W = \top_{D_I}$. Now define $\mathrm{Pl}_{\mathcal{P}}(U) = f_U$. Thus, $\mathrm{Pl}_{\mathcal{P}}(U) \leq \mathrm{Pl}_{\mathcal{P}}(V)$ iff $f_U(i) \leq f_V(i)$ for all $i \in I$ iff $\mu(U) \leq \mu(V)$ for all $\mu \in \mathcal{P}$. Clearly $\mathrm{Pl}_{\mathcal{P}}$ satisfies Pl1–3. Pl1 and Pl2 follow since $\mathrm{Pl}_{\mathcal{P}}(\emptyset) = f_\emptyset = \perp_{D_I}$ and $\mathrm{Pl}_{\mathcal{P}}(W) = f_W = \top_{D_I}$, while Pl3 follows since if $U \subseteq V$ then $\mu(U) \leq \mu(V)$ for all $\mu \in \mathcal{P}$. $\mathrm{Pl}_{\mathcal{P}}$ captures all the information in $\mathcal{P}$ (unlike, say, $\mathcal{P}_*$, which washes much of it away by taking infs).

This way of associating a plausibility measure with a set $\mathcal{P}$ of probability measures generalizes: it provides a way of associating a single plausibility measure with any set of plausibility measures; I leave the straightforward details to the reader.

**Possibility measures:** A *fuzzy measure* (or a *Sugeno measure*) $f$ on $W$ (Wang & Klir, 1992) is a function $f : 2^W \mapsto [0, 1]$, that satisfies Pl1–3. (That is, it is less general than a plausibility measure only in that it requires the range to be $[0, 1]$ rather than an arbitrary partially ordered set.) A *possibility measure* Poss on $W$ is a special case of a Sugeno measure; it is a function mapping subsets of $W$ to $[0, 1]$ such that $\mathrm{Poss}(W) = 1$, $\mathrm{Poss}(\emptyset) = 0$, and $\mathrm{Poss}(U) = \sup_{w \in U}(\mathrm{Poss}(\{w\}))$, so that $\mathrm{Poss}(U \cup V) = \max(\mathrm{Poss}(U), \mathrm{Poss}(V))$ (Dubois & Prade, 1990). Clearly a possibility measure is a plausibility measure.

**Ranking functions:** An *ordinal ranking* (or *$\kappa$-ranking* or *ranking function*) $\kappa$ on $W$ (as defined by (Goldszmidt & Pearl, 1992), based on ideas that go back to (Spohn, 1988)) is a function mapping subsets of $W$ to $I\!\!N^* = I\!\!N \cup \{\infty\}$ such that $\kappa(W) = 0$, $\kappa(\emptyset) = \infty$, and $\kappa(U) = \min_{w \in U}(\kappa(\{w\}))$, so that $\kappa(U \cup V) = \min(\kappa(U), \kappa(V))$. Intuitively, a ranking function assigns a degree of surprise to each subset of worlds in $W$, where 0 means unsurprising and higher numbers denote greater surprise. It is easy to see that if $\kappa$ is a ranking function on $W$, then $(W, 2^W, \kappa)$ is a plausibility space, where $x \leq_{I\!\!N^*} y$ if and only if $y \leq x$ under the usual ordering on the natural numbers. One standard view of a ranking function, going back to Spohn, is that a ranking of $k$ can be associated with a probability of $\epsilon^k$, for some fixed (possibly infinitesimal) $\epsilon$. Note that this viewpoint justifies taking $\kappa(W) = 0$, $\kappa(\emptyset) = \infty$, and $\kappa(U \cup V) = \min(\kappa(U), \kappa(V))$.

---

1. In the conference version of this paper (Halpern, 2000), $D_I$, the range of the plausibility measure, was taken to be functions from $\mathcal{P}$ to $[0, 1]$, not from the index set $I$ to $[0, 1]$. The difference is mainly cosmetic, but this representation makes the range independent of $\mathcal{P}$, so that the same plausibility values can be used for any set of probability measures indexed by $I$.

## 2.2 Conditional Plausibility Measures

Since Bayesian networks make such heavy use of conditioning, my interest here is not just plausibility measures, but *conditional* plausibility measures (*cpm*'s). Given a set $W$ of worlds, a cpm maps pairs of subsets of $W$ to some partially ordered set $D$. I write $\text{Pl}(U|V)$ rather than $\text{Pl}(U, V)$, in keeping with standard notation for conditioning. In the case of a probability measure $\mu$, it is standard to take $\mu(U|V)$ to be undefined in $\mu(V) = 0$. In general, we must make precise what the allowable second arguments are. Thus, I take the domain of a cpm to have the form $\mathcal{F} \times \mathcal{F}'$ where, intuitively, $\mathcal{F}'$ consists of those sets in $\mathcal{F}$ on which it makes sense to condition. For example, for a conditional probability measure defined in the usual way from an unconditional probability measure $\mu$, $\mathcal{F}'$ consists of all sets $V$ such that $\mu(V) > 0$. (Note that $\mathcal{F}'$ is not an algebra—it is not closed under complementation.) A *Popper algebra* over $W$ is a set $\mathcal{F} \times \mathcal{F}'$ of subsets of $W \times W$ satisfying the following properties:

Acc1. $\mathcal{F}$ is an algebra over $W$.

Acc2. $\mathcal{F}'$ is a nonempty subset of $\mathcal{F}$.

Acc3. $\mathcal{F}'$ is closed under supersets in $\mathcal{F}$; that is, if $V \in \mathcal{F}'$, $V \subseteq V'$, and $V' \in \mathcal{F}$, then $V' \in \mathcal{F}'$.

(Popper algebras are named after Karl Popper, who was the first to consider formally conditional probability as the basic notion (Popper, 1968). De Finetti (1936) also did some early work, apparently independently, taking conditional probabilities as primitive. Indeed, as Rényi (1964) points out, the idea seems to go back as far as Keynes (1921).)

A *conditional plausibility space* (*cps*) is a tuple $(W, \mathcal{F}, \mathcal{F}', \text{Pl})$, where $\mathcal{F} \times \mathcal{F}'$ is a Popper algebra over $W$, $\text{Pl} : \mathcal{F} \times \mathcal{F}' \to D$, $D$ is a partially ordered set of plausibility values, and $\text{Pl}$ is a *conditional plausibility measure* (cpm) that satisfies the following conditions:

CPl1. $\text{Pl}(\emptyset|V) = \bot_D$.

CPl2. $\text{Pl}(W|V) = \top_D$.

CPl3. If $U \subseteq U'$, then $\text{Pl}(U|V) \leq \text{Pl}(U'|V)$.

CPl4 $\text{Pl}(U|V) = \text{Pl}(U \cap V|V)$.

CPl1–3 are the obvious analogues to Pl1–3. CPl4 is a minimal property that guarantees that when conditioning on $V$, everything is relativized to $V$. It follows easily from CPl1–4 that $\text{Pl}(\cdot|V)$ is a plausibility measure on $V$ for each fixed $V$. A cps is *acceptable* if it satisfies

Acc4. If $V \in \mathcal{F}'$, $U \in \mathcal{F}$, and $\text{Pl}(U|V) \neq \bot_D$, then $U \cap V \in \mathcal{F}'$.

Acceptability is a generalization of the observation that if $\Pr(V) \neq 0$, then conditioning on $V$ should be defined. It says that if $\mathrm{Pl}(U|V) \neq \perp_D$, then conditioning on $V \cap U$ should be defined.

CPl1–4 are rather minimal requirements. For example, they do not place any constraints on the relationship between $\mathrm{Pl}(U|V)$ and $\mathrm{Pl}(U|V')$ if $V \neq V'$. One natural additional condition is the following.

CPl5. If $V \cap V' \in \mathcal{F}'$ and $U, U' \in \mathcal{F}$, then $\mathrm{Pl}(U|V \cap V') \leq \mathrm{Pl}(U'|V \cap V')$ iff $\mathrm{Pl}(U \cap V|V') \leq \mathrm{Pl}(U' \cap V|V')$.

It is not hard to show that CPl5 implies CPl4.

**Lemma 2.1:** *CPl5 implies CPl4.*

**Proof:** Since clearly $\mathrm{Pl}(U \cap V|V) = \mathrm{Pl}(U \cap V \cap V|V)$, by CPl5 it follows that $\mathrm{Pl}(U|V \cap V) = \mathrm{Pl}(U \cap V|V \cap V)$, and hence $\mathrm{Pl}(U|V) = \mathrm{Pl}(U \cap V|V)$. ∎

CPl5 does not follow from CPl1–4 (indeed, as shown below, the standard notion of conditioning for lower probabilities satisfies CPl1–4 but not CPl5). A cps that satisfies CPl5 is said to be *coherent*. Although I do not assume CPl5 here, it in fact holds for all plausibility measures to which one of the main results applies (see Lemma 3.5).

In any case, CPl5 is certainly not the only coherence that might be required. For example, it may seem reasonable to require that if $V$ and $V'$ are disjoint, then it is not the case that both $\mathrm{Pl}(U|V \cup V') > \mathrm{Pl}(U|V)$ and $\mathrm{Pl}(U|V \cup V') > \mathrm{Pl}(U|V')$. Similarly, we may want to require that it not be the case that $\mathrm{Pl}(U|V \cup V') < \mathrm{Pl}(U|V)$ and $\mathrm{Pl}(U|V \cup V') < \mathrm{Pl}(U|V')$.[2] Coming up with a reasonable set of coherence conditions remains a topic for future work. The only properties needed for the results of this paper are CPl1–4.

The notion of cps considered here is closely related to that defined in (Friedman & Halpern, 1995). There, a cps is taken to be a family $\{W, D_V, \mathrm{Pl}_V) : V \subseteq W, V \neq \emptyset\}$ of plausibility spaces, where each plausibility measure $\mathrm{Pl}_V$ satisfies Pl1–3 and has domain $2^W$ and an analogue of CPl5 holds: $\mathrm{Pl}_{V \cap V'}(U) \leq \mathrm{Pl}_{V \cap V'}(U')$ iff $\mathrm{Pl}_{V'}(U \cap V) \leq \mathrm{Pl}_{V'}(U' \cap V)$. To distinguish the definition of cps given in this paper from that given in (Friedman & Halpern, 1995), I call the latter an FH-cps. There is no analogue to Acc1–4 in (Friedman & Halpern, 1995); $\mathcal{F}$ is implicitly taken to be $2^W$, while $\mathcal{F}'$ is implicitly taken to be $2^W - \{\emptyset\}$. This is an inessential difference between the definitions. More significantly, note that in an FH-cps, $(W, D_V, \mathrm{Pl}_V)$ is a plausibility space for each fixed $V$, and thus satisfies Pl1–3. However, requiring CPl1–3 is *a priori* stronger than requiring Pl1–3 for each separate plausibility space. Pl1 requires that $\mathrm{Pl}(\emptyset|V) = \perp_{D_V}$, but the elements $\perp_{D_V}$ may be different for each $V$. By way of contrast, CPl1 requires that $\mathrm{Pl}(\perp|V)$ must be the same element, $\perp_D$, for all $V$. Similar remarks hold for Pl2. Nevertheless, as is shown below, there is a construction that converts an FH-cps to a coherent cps.

I now consider some standard ways of getting a cps starting with an unconditional representation of uncertainty.

**Definition 2.2:** A cps $(W, \mathcal{F}, \mathcal{F}', \mathrm{Pl})$ *extends* an unconditional plausibility space $(W, \mathcal{F}, \mathrm{Pl}')$ if $\mathrm{Pl}(U|W) = \mathrm{Pl}'(U)$. $(W, \mathcal{F}, \mathcal{F}', \mathrm{Pl})$ is *standard* if $\mathcal{F}' = \{U : \mathrm{Pl}(U) \neq \perp\}$. ∎

All the constructions below result in standard cps's.

---

2. I think an anonymous referee of this paper for suggesting this condition.

**Ranking functions:** Given an unconditional ranking function $\kappa$, there is a well-known way of extending it to a conditional ranking function:

$$\kappa(U|V) = \begin{cases} \kappa(U \cap V) - \kappa(V) & \text{if } \kappa(V) \neq \infty, \\ \text{undefined} & \text{if } \kappa(V) = \infty. \end{cases}$$

This is consistent with the view that if $\kappa(V) = k$, then $\mu(V) = \epsilon^k$, since then $\kappa(U|V) = \epsilon^{\kappa(U \cap V) - \kappa(V)}$. It is easy to check that this definition results in a coherent cps.

**Possibility measures:** There are two standard ways of defining a conditional possibility measure from an unconditional possibility measure Poss. To distinguish them, I write $\text{Poss}(U|V)$ for the first approach and $\text{Poss}(U||V)$ for the second approach. According to the first approach,

$$\text{Poss}(U|V) = \begin{cases} \text{Poss}(V \cap U) & \text{if } \text{Poss}(V \cap U) < \text{Poss}(V), \\ 1 & \text{if } \text{Poss}(V \cap U) = \text{Poss}(V) > 0, \\ \text{undefined} & \text{if } \text{Poss}(V) = 0. \end{cases}$$

The second approach looks more like conditioning in probability:

$$\text{Poss}(U||V) = \begin{cases} \text{Poss}(V \cap U)/\text{Poss}(V) & \text{if } \text{Poss}(V) > 0, \\ \text{undefined} & \text{if } \text{Poss}(V) = 0. \end{cases}$$

It is easy to show that both definitions result in a coherent cps. (Many other notions of conditioning for possibility measures can be defined; see, for example (Fonck, 1994). I focus on these two because they are the ones most-often considered in the literature.)

**Sets of probabilities:** For a set $\mathcal{P}$ of probabilities, conditioning can be defined for all the representations of $\mathcal{P}$ as a plausibility measure. But in each case there are subtle choices involving when conditioning is undefined. For example, one definition of conditional lower probability is that $\mathcal{P}_*(U|V) = \inf\{\mu(U|V) : \mu(V) \neq 0\}$ if $\mu(V) \neq 0$ for all $\mu \in \mathcal{P}$, and is undefined otherwise (i.e., if $\mu(V) = 0$ for some $\mu \in \mathcal{P}$). It is easy to check that $\mathcal{P}_*$ defined this way gives a coherent cpm, as does the corresponding definition of $\mathcal{P}^*$. The problem with this definition is that it may result in a rather small set $\mathcal{F}'$ for which conditioning is defined. For example, if for each set $V \neq W$, there is some measure $\mu \in \mathcal{P}$ such that $\mu(V) = 0$ (which can certainly happen in some nontrivial examples), then $\mathcal{F}' = \{W\}$. As a consequence, the cps defined in this way is not acceptable (i.e., does not satisfy Acc4) in general.

The following definition gives a lower probability which is defined on more arguments:

$$\mathcal{P}_*(U|V) = \begin{cases} \inf\{\mu(U|V) : \mu(V) \neq 0\} & \text{if } \mu(V) \neq 0 \text{ for some } \mu \in \mathcal{P}, \\ \text{undefined} & \text{if } \mu(V) = 0 \text{ for all } \mu \in \mathcal{P}. \end{cases}$$

It is easy to see that this definition agrees with the first one whenever the first is defined and results, in general, in a larger set $\mathcal{F}'$. Moreover, the resulting cps is acceptable. However, the second definition does not satisfy CPl5. For example, suppose that $W = \{a, b, c\}$ and $\mathcal{P} = \{\mu, \mu'\}$, where $\mu(a) = \mu(b) = 0$, $\mu(c) = 1$, $\mu'(a) = 2/3$, $\mu'(b) = 1/3$, and $\mu'(c) = 0$.

Taking $V = \{a, b\}$, $U = \{a\}$, and $U' = \{b\}$, it is easy to see that according to the second definition, $\mathcal{P}_*(U \cap V | W) = \mathcal{P}_*(U' \cap V | W) = 0$, but $\mathcal{P}_*(U|V) > \mathcal{P}_*(U'|V)$.

For $\mathrm{Pl}_{\mathcal{P}}$, there are two analogous definitions. For the first, $\mathrm{Pl}_{\mathcal{P}}(U|V)$ is defined only if $\mu(V) > 0$ for all $\mu \in \mathcal{P}$, in which case $\mathrm{Pl}_{\mathcal{P}}(U|V)$ is $f_{U|V}$, where $f_{U|V}(i) = \mu_i(U|V)$. This definition gives a coherent cps, but again, in general, not one that is acceptable. In this paper, I focus on the following definition, which does result in an acceptable cps.

First extend $D_I$ by allowing functions which have value $*$ (intuitively, $*$ denotes undefined). More precisely, let $D'_I$ consist of all functions $f$ from $I$ to $[0,1] \cup \{*\}$ such that $f(i) \neq *$ for at least one $i \in I$. The idea is to define $\mathrm{Pl}_{\mathcal{P}}(U|V) = f_{U|V}$, where $f_{U|V}(i) = \mu_i(U|V)$ if $\mu_i(V) > 0$ and $*$ otherwise. (Note that this agrees with the previous definition, which applies only to the situation where $\mu(V) > 0$ for all $\mu \in \mathcal{P}$.) There is a problem though, one to which I have already alluded. CP11 says that $f_{\emptyset|V}$ must be $\perp$ for all $V$. Thus, it must be the case that $f_{\emptyset|V_1} = f_{\emptyset|V_2}$ for all $V_1, V_2 \subseteq W$. But if $\mu_i \in \mathcal{P}$ and $V_1, V_2 \subseteq W$ are such that $\mu_i(V_1) > 0$ and $\mu_i(V_2) = 0$, then $f_{\emptyset|V_1}(i) = 0$ and $f_{\emptyset|V_2}(i) = *$, so $f_{\emptyset|V_1} \neq f_{\emptyset|V_2}$. A similar problem arises with CP12.

To deal with this problem $D'_I$ must be slightly modified. Say that $f \in D'_I$ is *equivalent to* $\perp_{D'_I}$ if $f(i)$ is either $0$ or $*$ for all $i \in I$; similarly, $f$ *is equivalent to* $\top_{D'_I}$ if $f(i)$ is either $1$ or $*$ for all $i \in I$. (Since, by definition, $f(i) \neq *$ for at least one $i \in I$, an element cannot be equivalent to both $\top_{D'_I}$ and $\perp_{D'_I}$.) Let $D^*_I$ be the same as $D'_I$ except that all elements equivalent to $\perp_{D_I}$ are identified (and viewed as one element) and all elements equivalent to $\top_{D_I}$ are identified. More precisely, let $D^*_I = \{\perp_{D^*_I}, \top_{D^*_I}\} \cup \{f \in D' : f$ is not equivalent to $\top_{D^*_I}$ or $\perp_{D^*_I}\}$. Define the ordering $\leq$ on $D^*_I$ by taking $f \leq g$ if one of the following three conditions holds:

- $f = \perp_{D^*_I}$,

- $g = \top_{D^*_I}$,

- neither $f$ nor $g$ is $\perp_{D^*_I}$ or $\top_{D^*_I}$ and for all $i \in I$, either $f(i) = g(i) = *$ or $f(i) \neq *$, $g(i) \neq *$, and $f(i) \leq g(i)$.

Now define

$$\mathrm{Pl}_{\mathcal{P}}(U|V) = \begin{cases} \perp_{D^*_I} & \text{if } \mu(V) \neq 0 \text{ for some } \mu \in \mathcal{P} \text{ and} \\ & \mu(V) \neq 0 \text{ implies } \mu(U|V) = 0 \text{ for all } \mu \in \mathcal{P}, \\ \top_{D^*_I} & \text{if } \exists \mu \in \mathcal{P}(\mu(V) \neq 0) \text{ and } \forall \mu \in \mathcal{P}(\mu(V) \neq 0 \Rightarrow \mu(U|V) = 1), \\ \text{undefined} & \text{if } \mu(V) = 0 \text{ for all } \mu \in \mathcal{P}, \\ f_{U|V} & \text{otherwise.} \end{cases}$$

It is easy to check that this gives a coherent cps.

**Plausibility measures:** The construction for $\mathrm{Pl}_{\mathcal{P}}$ can be used to convert any FH-cps to a cps. I demonstrate the idea by showing how to construct a conditional plausibility measure from an unconditional plausibility measure. Given an unconditional plausibility space $(W, \mathcal{F}, \mathrm{Pl})$ with range $D$, an FH-cps is constructed in (Friedman & Halpern, 1995) by defining $\mathrm{Pl}(U|V) = \mathrm{Pl}(U \cap V)$. Thus, $D_V = \{d \in D : d \leq \mathrm{Pl}(V)\}$ and $\top_{D_V} = \mathrm{Pl}(V)$. This is not a cps because CP12 is not satisfied, but it is an FH-cps, since Pl1–3 is satisfied for each fixed $V$, and so is CP15. As observed in (Friedman & Halpern, 1995), this is in fact the FH-cps extending Pl that makes the minimal number of comparisons, in the sense that if $\text{Pl}'$ is an FH-cps extending Pl and $\text{Pl}(U|V) \leq \text{Pl}(U'|V)$, then $\text{Pl}'(U|V) \leq \text{Pl}'(U'|V)$.

To get a cps, let $\mathcal{D}' = \{(d, V) : V \subseteq W, d \leq \text{Pl}(V), \text{Pl}(V) > \perp_D\}$. Say that $(d, V)$ is *equivalent to* $\perp_{D^*}$ if $d = \perp_D$; say that $(d, V)$ is *equivalent to* $\top_{D^*}$ if $d = \text{Pl}(V)$. Now let $D^* = \{\perp_{D^*}, \top_{D^*}\} \cup \{f \in D' : f \text{ is not equivalent to } \top_{D^*} \text{ or } \perp_{D^*}\}$. Then define $d \leq_{D^*} d'$ for $d, d' \in D^*$ iff $d = \perp_{D^*}$, $d' = \top_{D^*}$, or there is some $V \subseteq W$ such that $d = (d_1, V)$, $d' = (d_2, V)$, and $d_1 \leq_D d_2$. Finally, for $U, V \in \mathcal{F}$, define

$$\text{Pl}(U|V) = \begin{cases} (\text{Pl}(U \cap V), V) & \text{if } \perp_D < \text{Pl}(U \cap V) < \text{Pl}(V), \\ \top_{D^*} & \text{if } \text{Pl}(U \cap V) = \text{Pl}(V) > \perp_D, \\ \perp_{D^*} & \text{if } \text{Pl}(U \cap V) = \perp_D, \text{Pl}(V) > \perp_D, \\ \text{undefined} & \text{if } \text{Pl}(V) = \perp_D. \end{cases}$$

I leave it to the reader to check that Pl is a coherent cpm. It is important that $\text{Pl}(U|V)$ is undefined if $\text{Pl}(V) = \perp_D$; if we tried to extend the construction to $V$ such that $\text{Pl}(V) = \perp_D$, then we would have $\top_{D^*} = \perp_{D^*}$. This issue did not arise in (Friedman & Halpern, 1995), since there were separate plausibility spaces for each choice of $V$.

## 3. Algebraic Conditional Plausibility Measures

To be able to carry out the type of reasoning used in Bayesian networks, it does not suffice to just have conditional plausibility. We need to have analogues of addition and multiplication. More precisely, there needs to be some way of computing the plausibility of the union of two disjoint sets in terms of the plausibility of the individual sets and a way of computing $\text{Pl}(U \cap V|V')$ given $\text{Pl}(U|V \cap V')$ and $\text{Pl}(V|V')$.

**Definition 3.1:** A cps $(W, \mathcal{F}, \mathcal{F}', \text{Pl})$ where Pl has range $D$ is *algebraic* if it is acceptable and there are functions $\oplus : D \times D \to D$ and $\otimes : D \times D \to D$ such that the following properties hold:

Alg1. If $U, U' \in \mathcal{F}$ are disjoint and $V \in \mathcal{F}'$ then $\text{Pl}(U \cup U'|V) = \text{Pl}(U|V) \oplus \text{Pl}(U'|V)$.

Alg2. If $U \in \mathcal{F}$, $V \cap V' \in \mathcal{F}'$, then $\text{Pl}(U \cap V|V') = \text{Pl}(U|V \cap V') \otimes \text{Pl}(V|V')$.

Alg3. $\otimes$ distributes over $\oplus$; more precisely, $a \otimes (b_1 \oplus \cdots \oplus b_n) = (a \otimes b_1) \oplus \cdots \oplus (a \otimes b_n)$ if $(a, b_1), \ldots, (a, b_n), (a, b_1 \oplus \cdots \oplus b_n) \in Dom_{\text{Pl}}(\otimes)$ and $(b_1, \ldots, b_n), (a \otimes b_1, \ldots, a \otimes b_n) \in Dom_{\text{Pl}}(\oplus)$, where $Dom_{\text{Pl}}(\oplus) = \{(\text{Pl}(U_1|V), \ldots, \text{Pl}(U_n|V)) : U_1, \ldots, U_n \in \mathcal{F} \text{ are pairwise disjoint and } V \in \mathcal{F}'\}$ and $Dom_{\text{Pl}}(\otimes) = \{(\text{Pl}(U|V \cap V'), \text{Pl}(V|V')) : U \in \mathcal{F}, V \cap V' \in F'\}$.[3] (See below for a discussion of $Dom_{\text{Pl}}(\oplus)$ and $Dom_{\text{Pl}}(\otimes)$. In the sequel, I omit the subscript Pl if it is clear from context.)

Alg4. If $(a, c), (b, c) \in Dom(\otimes)$, $a \otimes c \leq b \otimes c$, and $c \neq \perp$, then $a \leq b$.

---

3. In the conference version of this paper, $Dom(\oplus)$ was taken to consist only of pairs, not tuples of arbitrary finite length, and distributivity was considered only for terms of the form $a \otimes (b \oplus b')$. The more general version considered here is slightly stronger. The reason is that it is possible that $(a, b_1 \oplus \cdots \oplus b_n) \in Dom(\otimes)$ even though $(a, b_1 \oplus \cdots \oplus b_k) \notin Dom(\otimes)$ for $k \leq n$. Note also that only left distributivity is required here.

I sometimes refer to the cpm Pl as being algebraic as well. ∎

It may seem more natural to consider a stronger version of Alg4 that applies to all pairs in $D \times D$, such as

Alg4′. If $a \otimes c \leq b \otimes c$ and $c \neq \perp$, then $a \leq b$.

However, as Proposition 3.2 below shows, by requiring that Alg3 and Alg4 hold only for tuples in $Dom(\oplus)$ and $Dom(\otimes)$ rather than on all tuples in $D \times D$, some cps's of interest become algebraic that would otherwise not be. Intuitively, we care about $\otimes$ mainly to the extent that Alg1 and Alg2 holds, and Alg1 and Alg2 apply only to tuples in $Dom(\oplus)$ and $Dom(\otimes)$, respectively. Thus, it does not seem unreasonable that Alg4 be required to hold only for these tuples.

**Proposition 3.2:** *The constructions for extending an unconditional probability measure, ranking function, possibility measure (using either* $\mathrm{Poss}(U|V)$ *or* $\mathrm{Poss}(U||V)$*), and the plausibility measure* $Pl_{\mathcal{P}}$ *defined by a set* $\mathcal{P}$ *of probability measures to a cps result in algebraic cps's.*[4]

**Proof:** It is easy to see that in each case the cps is acceptable. It is also easy to find appropriate notions of $\otimes$ and $\oplus$ in the case of probability measures, ranking functions, and possibility measures using $\mathrm{Poss}(U||V)$. For probability, clearly $\oplus$ and $\otimes$ are essentially $+$ and $\times$; however, since the range of probability is $[0,1]$, $a \oplus b$ must be defined as $\max(1, a+b)$, and Alg3 holds only for $Dom(\oplus) = \{(a_1, \ldots, a_k) : a_1 + \cdots + a_k \leq 1\}$; there is no constraint on $Dom(\times)$; it is $[0,1] \times [0,1]$. For ranking, $\oplus$ and $\otimes$ are min and $+$; there are no constraints on $Dom(\min)$ and $Dom(+)$. For $\mathrm{Poss}(U||V)$, $\oplus$ is max and $\otimes$ is $\times$; again, there are no constraints on $Dom(\max)$ and $Dom(\times)$. I leave it to the reader to check that Alg1–4 hold in all these cases.

For $\mathrm{Poss}(U|V)$, $\oplus$ is again max and $\otimes$ is min. There are no constraints on $Dom(\max)$; however, note that $(a, b) \in Dom(\min)$ iff either $a < b$ or $a = 1$. For suppose that $(a, b) = (\mathrm{Poss}(U|V \cap V'), \mathrm{Poss}(V|V'))$, where $U \in \mathcal{F}$ and $V \cap V' \in \mathcal{F}'$. If $\mathrm{Poss}(U \cap V \cap V') = \mathrm{Poss}(V \cap V')$ then $a = \mathrm{Poss}(U|V \cap V') = 1$; otherwise, $\mathrm{Poss}(U \cap V \cap V') < \mathrm{Poss}(V \cap V')$, in which case $a = \mathrm{Poss}(U|V \cap V') = \mathrm{Poss}(U \cap V \cap V') < \mathrm{Poss}(V \cap V') \leq \mathrm{Poss}(V|V') = b$. It is easy to check Alg1–3. While min does not satisfy Alg4′—certainly $\min(a, c) = \min(b, c)$ does not in general imply that $a = b$—Alg4 does hold. For if $\min(a, c) \leq \min(b, c)$ and $a = 1$, then clearly $b = 1$. Alternatively, if $a < c$, then $\min(a, c) = a$ and the only way that $a \leq \min(b, c)$, given that $b < c$ or $b = 1$, is if $a \leq b$.

Finally, for $Pl_{\mathcal{P}}$, $\oplus$ and $\otimes$ are essentially pointwise addition and multiplication. But there are a few subtleties. As in the case of probability, $Dom(\oplus)$ consists of sequences which sum to at most 1 for each index $i$. Care must also be taken in dealing with $\perp_{D_I^*}$ and $\top_{D_I^*}$. More precisely, $Dom(\oplus)$ consists of all tuples $(f_1, \ldots, f_n)$ such that either

1(a). $f_j \neq \top_{D_I^*}, j = 1, \ldots, n$,

1(b). if $f_j, f_k \neq \perp_{D_I^*}$ for $1 \leq j, k \leq n$, then $f_j(i) = *$ iff $f_k(i) = *$, for all $i \in I$, and

---

4. Essentially the same result is proved in (Friedman & Halpern, 1995) for all cases but $Pl_{\mathcal{P}}$.

1(c). $\sum_{\{j: f_j \neq \perp_{D_I^*}, f_j(i) \neq *\}} f_j(i) \leq 1$

or

2. there exists $j$ such that $f_j = \top_{D_I^*}$ and $f_k = \perp_{D_I^*}$ for $k \neq j$;

$Dom(\otimes)$ consists of pairs $(f, g)$ such that either one of $f$ or $g$ is in $\{\perp_{D_I^*}, \top_{D_I^*}\}$ or neither $f$ nor $g$ is in $\{\perp_{D_I^*}, \top_{D_I^*}\}$ and $g(i) \in \{0, *\}$ iff $f(i) = *$. The definition of $\oplus$ is relatively straightforward. Define $f \oplus \top_{D_I^*} = \top_{D_I^*} \oplus f = \top_{D_I^*}$ and $f \oplus \perp_{D_I^*} = \perp_{D_I^*} \oplus f = f$. If $f, g \cap \{\perp_{D_I^*}, \top_{D_I^*}\} = \emptyset$, then $f \oplus g = h$, where $h(i) = \min(1, f(i) + g(i))$ (taking $a + * = * + a = *$ and $\min(1, *) = *$). In a similar spirit, define $f \otimes \top_{D_I^*} = \top_{D_I^*} \otimes f = f$ and $f \otimes \perp_{D_I^*} = \perp_{D_I^*} \otimes f = \perp_{D_I^*}$; if $\{f, g\} \cap \{\perp_{D_I^*}, \top_{D_I^*}\} = \emptyset$, then $f \otimes g = h$, where $h(i) = f(i) \times g(i)$ (taking $* \times a = a \times * = *$ if $a \neq 0$ and $* \times 0 = 0 \times * = 0$). It is important that $* \times 0 = 0$ and $* \times * = *$, since otherwise Alg3 may not hold. For example, according to Alg3,

$$((1/2, *, 1/2) \otimes (a, 0, b)) \oplus ((1/2, *, 1/2)) \otimes (a, 0, b)) = ((1/2, *, 1/2) \oplus (1/2, *, 1/2)) \otimes (a, 0, b) = (a, 0, b)$$

(since $(1/2, *, 1/2) \oplus (1/2, *, 1/2) = \top_{D_I^*}$) and, similarly, $((1/2, *, 1/2) \otimes (a, *, b)) \oplus ((1/2, *, 1/2)) \otimes (a, *, b)) = (a, *, b)$. Since $* \times 0 = 0$ and $* \times * = *$, these equalities hold. I leave it to the reader to check that, with these definitions, Alg1–4 hold (although note that the restrictions to $Dom(\oplus)$ and $Dom(\otimes)$ are required for both Alg3 and Alg4 to hold). ∎

Conditional lower probability is not algebraic. For example, it is not hard to construct pairwise disjoint sets $U_1$, $V_1$, $U_2$, and $V_2$ and a set $\mathcal{P}$ of probability measures such that $\mathcal{P}_*(U_i) = \mathcal{P}_*(V_i)$ (and $\mathcal{P}^*(U_i) = \mathcal{P}^*(V_i)$) for $i = 1, 2$, but $\mathcal{P}_*(U_1 \cup U_2) \neq \mathcal{P}_*(V_1 \cup V_2)$. That means there cannot be a function $\oplus$ in the case of lower probability.[5]

For later convenience, I list some simple properties of algebraic cpms that show that $\perp$ and $\top$ act like 0 and 1 with respect to addition and multiplication. Let $Range(\mathrm{Pl}) = \{d : \mathrm{Pl}(U|V) = d$ for some $(U, V) \in \mathcal{F} \times \mathcal{F}'\}$.

**Lemma 3.3:** *If $(W, \mathcal{F}, \mathcal{F}', Pl)$ is an algebraic cps, then $d \oplus \perp = \perp \oplus d = d$ for all $d \in Range(Pl)$.*

**Proof:** Suppose that $d = \mathrm{Pl}(U|V)$. By Alg1, it follows that

$$d = \mathrm{Pl}(U|V) = \mathrm{Pl}(U \cup \emptyset | V) = \mathrm{Pl}(U|V) \oplus \mathrm{Pl}(\emptyset | V) = d \oplus \perp.$$

A similar argument shows that $d = \perp \oplus d$. ∎

**Lemma 3.4:** *If $(W, \mathcal{F}, \mathcal{F}', Pl)$ is an algebraic cps then, for all $d \in Range(Pl)$,*

(a) $d \otimes \top = d;$

---

5. For readers familiar with Dempster-Shafer belief functions (Shafer, 1976), they provide another example of a plausibility measure. There are two well-known ways of defining conditioning for belief functions (see (Fagin & Halpern, 1991)), one using Dempster's rule of combination and the other treating belief functions as lower probabilities. Neither leads to an algebraic cps, which is why I have not discussed belief functions in this paper.

*(b) if $d \neq \bot$, then $\top \otimes d = d$;*

*(c) if $d \neq \bot$, then $\bot \otimes d = \bot$;*

*(d) if $(d, \bot) \in Dom(\otimes)$, then $\top \otimes \bot = d \otimes \bot = \bot \otimes \bot = \bot$.*

**Proof:** Suppose that $d = \text{Pl}(U|V)$. By Alg2, CPl2, and CPl4, it follows that

$$d = \text{Pl}(U|V) = \text{Pl}(U \cap V|V) = \text{Pl}(U|V) \otimes \text{Pl}(V|V) = d \otimes \top.$$

Similarly, if $d \neq \bot$, then $U \cap V \in \mathcal{F}'$ (by Acc4), so

$$d = \text{Pl}(U|V) = \text{Pl}(U \cap V|V) = \text{Pl}(U \cap V|U \cap V) \otimes \text{Pl}(U \cap V|V) = \top \otimes d.$$

If $d \neq \bot$, then by Alg2, CPl1, and CPl4

$$\bot = \text{Pl}(\bot|V) = \text{Pl}(\bot|U \cap V) \otimes \text{Pl}(U|V) = \bot \otimes d.$$

Finally, if $(d, \bot) \in Dom(\otimes)$, then there exist $U, V, V'$ such that $V \cap V' \in \mathcal{F}'$, $\text{Pl}(U|V \cap V') = d$ and $\text{Pl}(V|V') = \bot$. By Alg2, $\text{Pl}(U \cap V|V') = \text{Pl}(U|V \cap V') \otimes \text{Pl}(V|V') = d \otimes \bot$. By CPl3, $\text{Pl}(U \cap V|V') \leq \text{Pl}(V|V') = \bot$, so $\text{Pl}(U \cap V|V') = \bot$. Thus, $d \otimes \bot = \bot$. Replacing $U$ with $V \cap V'$, the same argument shows that $\top \otimes \bot = \bot$; replacing $U$ with $\emptyset$, we get that $\bot \otimes \bot = \bot$. ∎

I conclude this section by showing that a standard algebraic cps that satisfies one other minimal property must also satisfy CPl5. Say that $\otimes$ is *monotonic* if $d \leq d'$ and $e \leq e'$ then $d \otimes e \leq d' \otimes e'$. A cpm (cps) is monotonic if $\otimes$ is.

**Lemma 3.5:** *A standard algebraic monotonic cps satisfies CPl5.*

**Proof:** Suppose that $(W, \mathcal{F}, \mathcal{F}', \text{Pl})$ is a standard algebraic cps and that $V \cap V' \in \mathcal{F}'$. If $\text{Pl}(U|V \cap V') \leq \text{Pl}(U'|V \cap V')$, then it follows from Alg2 and monotonicity that

$$\text{Pl}(U \cap V|V') = \text{Pl}(U|V \cap V') \otimes \text{Pl}(V|V') \leq \text{Pl}(U'|V \cap V') \otimes \text{Pl}(V|V') = \text{Pl}(U' \cap V|V').$$

For the opposite implication, suppose that $\text{Pl}(U \cap V|V') \leq \text{Pl}(U' \cap V|V')$. Then, by Alg2,

$$\text{Pl}(U|V \cap V') \otimes \text{Pl}(V|V') \leq \text{Pl}(U'|V \cap V') \otimes \text{Pl}(V|V'). \tag{1}$$

Since $V \cap V' \in \mathcal{F}'$ and the cps is standard, it must be the case that $\text{Pl}(V \cap V') \neq \bot$. Hence (by CPl3), $\text{Pl}(V') \neq \bot$; moreover, $\text{Pl}(V|V') \neq \bot$ (otherwise $\text{Pl}(V \cap V') = \text{Pl}(V|V') \otimes \text{Pl}(V') = \bot$). Thus, by applying Alg4 to (1), it follows that $\text{Pl}(U|V \cap V') \leq \text{Pl}(U'|V \cap V')$. ∎

## 4. Independence

How can we capture formally the notion that two events are *independent*? Intuitively, it means that they have nothing to do with each other—they are totally unrelated; the occurrence of one has no influence on the other. None of the representations of uncertainty that we have been considering can express the notion of "unrelatedness" (whatever it might mean) directly. The best we can do is to capture the "footprint" of independence on the notion. For example, in the case of probability, if $U$ and $V$ are unrelated, it seems reasonable to expect that learning $U$ should not affect the probability of $V$ and symmetrically, learning $V$ should not affect the probability of $U$. "Unrelatedness" is, after all, a symmetric notion.[6] The fact that $U$ and $V$ are probabilistically independent (with respect to probability measure $\mu$) can thus be expressed as $\mu(U|V) = \mu(U)$ and $\mu(V|U) = \mu(V)$. There is a technical problem with this definition: What happens if $\mu(V) = 0$? In that case $\mu(U|V)$ is undefined. Similarly, if $\mu(U) = 0$ then $\mu(V|U)$ is undefined. It is conventional to say that, in this case, $U$ and $V$ are still independent. This leads to the following formal definition.

**Definition 4.1:** $U$ and $V$ are *probabilistically independent (with respect to probability measure $\mu$)* if $\mu(V) \neq 0$ implies $\mu(U|V) = \mu(U)$ and $\mu(U) \neq 0$ implies $\mu(V|U) = \mu(V)$. ∎

This does not look like the standard definition of independence in texts, but an easy calculation shows that it is equivalent.

**Proposition 4.2:** *The following are equivalent:*

(a) $\mu(U) \neq 0$ *implies* $\mu(V|U) = \mu(V)$,

(b) $\mu(U \cap V) = \mu(U)\mu(V)$,

(c) $\mu(V) \neq 0$ *implies* $\mu(U|V) = \mu(U)$.

Thus, in the case of probability, it would be equivalent to say that $U$ and $V$ are independent with respect to $\mu$ if $\mu(U \cap V) = \mu(U)\mu(V)$ or to require only that $\mu(U|V) = \mu(U)$ if $\mu(V) \neq 0$ without requiring that $\mu(V|U) = \mu(V)$ if $\mu(U) \neq 0$. However, these equivalences do not necessarily hold for other representations of uncertainty. The definition of independence I have given here seems to generalize more appropriately.[7]

The definition of probabilistic conditional independence is analogous.

**Definition 4.3:** $U$ and $V$ are *probabilistically independent given $V'$ (with respect to probability measure $\mu$)* if $\mu(V \cap V') \neq 0$ implies $\mu(U|V \cap V') = \mu(U|V')$ and $\mu(U \cap V') \neq 0$ implies $\mu(V|U \cap V') = \mu(V|V')$. ∎

It is immediate that $U$ and $V$ are (probabilistically) independent iff they are independent conditional on $W$.

The generalization to conditional plausibility measures (and hence to all other representations of uncertainty that we have been considering) is straightforward.

**Definition 4.4:** Given a cps $(W, \mathcal{F}, \mathcal{F}', \text{Pl})$, $U, V \in \mathcal{F}$ are *plausibilistically independent given $V' \in \mathcal{F}$ (with respect to the cpm Pl)*, written $I_{\text{Pl}}(U, V|V')$, if $V \cap V' \in \mathcal{F}'$ implies $\text{Pl}(U|V \cap V') = \text{Pl}(U|V')$ and $U \cap V' \in \mathcal{F}'$ implies $\text{Pl}(V|U \cap V') = \text{Pl}(V|V')$. ∎

---

6. Walley (1991) calls the asymmetric notion *irrelevance* and defines $U$ being independent of $V$ as $U$ is irrelevant to $V$ and $V$ is irrelevant to $U$. Although my focus here is independence, irrelevance is an interesting notion in its own right; see (Cozman, 1998; Cozman & Walley, 1999).

7. Another property of probabilistic independence is that if $U$ is independent of $V$ then $\overline{U}$ is independent of $V$. This too does not follow for the other representations of uncertainty, and Walley (1991) actually makes this part of his definition. Adding this requirement would not affect any of the results here, although it would make the proofs somewhat lengthier, so I have not made it part of the definition.

We are interested in conditional independence of random variables as well as in conditional independence of events. All the standard definitions extend to plausibility in a straightforward way. A *random variable* $X$ on $W$ is a function from $W$ to the reals. Let $\mathcal{R}(X)$ be the set of possible values for $X$ (that is, the set of values over which $X$ *ranges*). As usual, $X = x$ is the event $\{w : X(w) = x\}$. If $\mathbf{X} = \{X_1, \ldots, X_k\}$ is a set of random variables and $\mathbf{x} = (x_1, \ldots, x_k)$, let $\mathbf{X} = \mathbf{x}$ be an abbreviation for the event $X_1 = x_1 \cap \ldots \cap X_k = x_k$. A random variable is *measurable* with respect to cps $(W, \mathcal{F}, \mathcal{F}', \mathrm{Pl})$ if $X = x \in \mathcal{F}$ for all $x \in \mathcal{R}(X)$. For the rest of the paper, I assume that all random variables $X$ are measurable and that $\mathcal{R}(X)$ is finite for all random variables $X$. Random variables $X$ and $Y$ are independent with respect to plausibility measure $\mathrm{Pl}$ if the events $X = x$ and $Y = y$ are independent for all $x \in \mathcal{R}(X)$ and $y \in \mathcal{R}(Y)$. More generally, given sets $\mathbf{X}$, $\mathbf{Y}$, and $\mathbf{Z}$ of random variables, $\mathbf{X}$ and $\mathbf{Y}$ are plausibilistically independent given $\mathbf{Z}$ (with respect to $\mathrm{Pl}$), denoted $I_{\mathrm{Pl}}^{rv}(\mathbf{X}, \mathbf{Y}|\mathbf{Z})$, if $I_{\mathrm{Pl}}(\mathbf{X} = \mathbf{x}, \mathbf{Y} = \mathbf{x}|\mathbf{Z} = \mathbf{z})$ for all $\mathbf{x}$, $\mathbf{y}$, and $\mathbf{z}$. (Note that I am using $I_{\mathrm{Pl}}$ for conditional independence of events and $I_{\mathrm{Pl}}^{rv}$ for conditional independence of random variables.) If $\mathbf{Z} = \emptyset$, then $I_{\mathrm{Pl}}^{rv}(\mathbf{X}, \mathbf{Y}|\mathbf{Z})$ if $\mathbf{X}$ and $\mathbf{Y}$ are unconditionally independent, that is, if $I_{\mathrm{Pl}}(\mathbf{X} = \mathbf{x}, \mathbf{Y} = \mathbf{x}|W)$ for all $\mathbf{x}$, $\mathbf{y}$; if either $\mathbf{X} = \emptyset$ or $\mathbf{Y} = \emptyset$, then $I_{\mathrm{Pl}}^{rv}(\mathbf{X}, \mathbf{Y}|\mathbf{Z})$ is taken to be vacuously true.

Now consider the following four properties of random variables, called the *semi-graphoid properties* (Pearl, 1988), where $\mathbf{X}$, $\mathbf{Y}$, and $\mathbf{Z}$ are pairwise disjoint sets of variables.

CIRV1. If $I_{\mathrm{Pl}}^{rv}(\mathbf{X}, \mathbf{Y}|\mathbf{Z})$ then $I_{\mathrm{Pl}}^{rv}(\mathbf{Y}, \mathbf{X}|\mathbf{Z})$.

CIRV2. If $I_{\mathrm{Pl}}^{rv}(\mathbf{X}, \mathbf{Y} \cup \mathbf{Y}'|\mathbf{Z})$ then $I_{\mathrm{Pl}}^{rv}(\mathbf{X}, \mathbf{Y}|\mathbf{Z})$.

CIRV3. If $I_{\mathrm{Pl}}^{rv}(\mathbf{X}, \mathbf{Y} \cup \mathbf{Y}'|\mathbf{Z})$ then $I_{\mathrm{Pl}}^{rv}(\mathbf{X}, \mathbf{Y}|\mathbf{Y}' \cup \mathbf{Z})$.

CIRV4. If $I_{\mathrm{Pl}}^{rv}(\mathbf{X}, \mathbf{Y}|\mathbf{Z})$ and $I_{\mathrm{Pl}}^{rv}(\mathbf{X}, \mathbf{Y}'|\mathbf{Y} \cup \mathbf{Z})$ then $I_{\mathrm{Pl}}^{rv}(\mathbf{X}, \mathbf{Y} \cup \mathbf{Y}'|\mathbf{Z})$.

It is well known that CIRV1–4 hold for probability measures. The following result generalizes this. The proof is not difficult, although care must be taken to show that the result depends only on the properties of algebraic cpms.

**Theorem 4.5:** *CIRV1–4 hold for all algebraic cps's.*

**Proof:** See the appendix. ∎

Theorem 4.5, of course, is very dependent on the definition of conditional independence given here. Other notions of independence have been studied in the literature for specific representations of uncertainty. Perhaps the most common definition tries to generalize the observation that if $U$ and $V$ are probabilistically independent, then $\mu(U \cap V) = \mu(U) \times \mu(V)$. Zadeh (1978) considered this approach in the context of possibility measures, calling it *noninteraction*, but it clearly makes sense for any algebraic cpm.

**Definition 4.6:** $U$ *and* $V$ *do not interact given* $V'$ *(with respect to the algebraic cpm Pl)*, denoted $NI_{\mathrm{Pl}}(U, V|V')$, if $V' \in \mathcal{F}'$ implies that $\mathrm{Pl}(U \cap V|V') = \mathrm{Pl}(U|V') \otimes \mathrm{Pl}(V|V')$.[8] ∎

---

8. Shenoy (1994) defines a notion similar in spirit to noninteraction for random variables.

Fonck (1994) shows that noninteraction is strictly weaker than independence for a number of notions of independence for possibility measures. The following result shows that independence implies noninteraction for all algebraic cpms.

**Lemma 4.7:** *If $(W, \mathcal{F}, \mathcal{F}', Pl)$ is an algebraic cps, then $I_{\text{Pl}}(U, V | V')$ implies $NI_{\text{Pl}}(U, V | V')$.*

**Proof:** Suppose that $V' \in \mathcal{F}'$ and $I_{\text{Pl}}(U, V | V')$ holds. If $V \cap V' \in \mathcal{F}'$ then, from Alg2, it follows that

$$\text{Pl}(U \cap V | V') = \text{Pl}(U | V \cap V') \otimes \text{Pl}(V | V') = \text{Pl}(U | V') \otimes \text{Pl}(V | V').$$

On the other hand, if $V \cap V' \notin \mathcal{F}'$, then by Acc4, $\text{Pl}(V | V') = \bot$. By CPl3, $\text{Pl}(U \cap V | V') = \bot$, and by Lemma 3.3, $\text{Pl}(U | V') \otimes \text{Pl}(V | V') = \bot$. Thus, $\text{Pl}(U \cap V | V') = \text{Pl}(U | V') \otimes \text{Pl}(V | V')$. ∎

What about the converse to Lemma 4.7? The results of Fonck show that it does not hold in general—indeed, it does not hold for $\text{Poss}(U | V)$. So what is required for noninteraction to imply independence? The following lemma provides a sufficient condition.

**Lemma 4.8:** *If $(W, \mathcal{F}, \mathcal{F}', Pl)$ is a standard algebraic cps that satisfies Alg4′, then $NI_{\text{Pl}}(U, V | V')$ implies $I_{\text{Pl}}(U, V | V')$.*

**Proof:** Suppose that $V \cap V' \in \mathcal{F}'$ and $NI_{\text{Pl}}(U, V | V')$. Then by Alg2,

$$\text{Pl}(U \cap V | V') = \text{Pl}(U | V \cap V') \otimes \text{Pl}(V | V'). \tag{2}$$

By Acc3, $V' \in \mathcal{F}'$, so $NI_{\text{Pl}}(U, V | V')$ implies

$$\text{Pl}(U \cap V | V') = \text{Pl}(U | V') \otimes \text{Pl}(V | V'). \tag{3}$$

Since $V \cap V' \in \mathcal{F}'$ and $(W, \mathcal{F}, \mathcal{F}', \text{Pl})$ is standard, $\text{Pl}(V \cap V') \neq \bot$. Since $\text{Pl}(V \cap V') = \text{Pl}(V | V') \otimes \text{Pl}(V')$, it follows from Lemma 3.4 that $\text{Pl}(V | V') \neq \bot$. So, by Alg4′, (2), and (3), it follows that $\text{Pl}(U | V \cap V') = \text{Pl}(U | V')$. An identical argument shows that $\text{Pl}(V | U \cap V') = \text{Pl}(V | V')$ if $U \cap V' \in \mathcal{F}'$. Thus, $\mathcal{I}_{\text{Pl}}(U, V | V')$. ∎

Lemmas 4.7 and 4.8 show why noninteraction and independence coincide for conditional probability defined from unconditional probability, ranking functions, and possibility measures using $\text{Poss}(U \| V)$. Moreover, they suggest why they do not coincide in general. Since neither $\text{Poss}(U | V)$ nor $\text{Pl}_{\mathcal{P}}$ satisfy Alg4′, it is perhaps not surprising that in neither case does noninteraction imply conditional independence. (We shall shortly see an example in the case of $\text{Pl}_{\mathcal{P}}$; Fonck (1994) gives examples in the case of $\text{Poss}(U | V)$.) Indeed, noninteraction may not even imply conditional independence for an arbitrary conditional probability measure, as the following example shows.

**Example 4.9:** Suppose that $W = \{a, b\}$, $\mathcal{F} = 2^W$, $\mathcal{F}' = \mathcal{F} - \{\emptyset\}$, $\mu(a) = 1$, $\mu(b) = 0$, but $\mu(b | b) = 1$. It is easy to see that $\{b\}$ is not independent of itself, but $\{b\}$ does not interact with $\{b\}$, since $\mu(b) = \mu(b) \times \mu(b)$. Nevertheless, it is not hard to check that this conditional probability measure $\mu$ is algebraic and, in fact, satisfies Alg4′. However, it is not standard, since $\{b\} \in \mathcal{F}'$ although $\mu(b) = 0$. ∎

It is easy to see that the assumption of standardness is necessary in Lemma 4.8. For suppose that $(W, \mathcal{F}, \mathcal{F}', \text{Pl})$ is an arbitrary nonstandard algebraic cps for which $\top \neq \bot$. Since $(W, \mathcal{F}, \mathcal{F}', \text{Pl})$ is nonstandard, there must exist some $U \in \mathcal{F}'$ such that $\text{Pl}(U|W) = \bot$. But then

$$\bot = \text{Pl}(\emptyset|W) = \text{Pl}(\emptyset|U) \otimes \text{Pl}(U|W) = \bot \otimes \bot.$$

Thus

$$\text{Pl}(U|W) = \bot = \bot \otimes \bot = \text{Pl}(U|W) \otimes \text{Pl}(U|W),$$

so $NI_{\text{Pl}}(U, U|W)$. But $\text{Pl}(U|U) = \top \neq \bot = \text{Pl}(U)$, so $I_{\text{Pl}}(U, U|W)$ does not hold.

In general, Theorem 4.5 does not hold if we use $NI_{\text{Pl}}$ rather than $I_{\text{Pl}}$. That is, Alg1–4 do not suffice to ensure that CIRV1–4 hold for $NI_{\text{Pl}}$. Besides noninteraction, a number of different approaches to defining independence for possibility measures (Campos & Huete, 1999a, 1999b; Dubois, Fariñas del Cerro, Herzig, & Prade, 1994; Fonck, 1994) and for sets of probability measures (Campos & Huete, 1993; Campos & Moral, 1995; Cousa et al., 1999) have been considered. In general, Theorem 4.5 does not hold for them either. It is beyond the scope of this paper to discuss and compare these approaches to that considered here, but it is instructive to consider independence for sets of probability measures in a little more detail, especially for the representation $\text{Pl}_{\mathcal{P}}$.

De Campos and Moral (1995) define what the call *type-1* independence. $U$ and $V$ are *type-1 independent conditional on* $V'$ *with respect to* $\mathcal{P}$ if $U$ and $V$ are independent conditional on $V'$ with respect to every $\mu \in \mathcal{P}$. It is easy to check that type-1 independence is equivalent to noninteraction in the context of sets of probability measures. Thus, by Lemma 4.7, $I_{\text{Pl}_{\mathcal{P}}}(U, V|V')$ implies that $U$ and $V$ are type-1 independent conditional on $V'$ (and similarly for random variables). However, the converse does not necessarily hold, because the two approaches treat conditioning on events that have probability 0 according to some (but not all) of the measures in $\mathcal{P}$ differently. To see this, consider an example discussed by de Campos and Moral. Suppose a coin is known to be either double-headed or double-tailed and is tossed twice. This can be represented by $\mathcal{P} = \{\mu_0, \mu_1\}$, where $\mu_0(hh) = 1$ and $\mu_0(ht) = \mu_0(th) = \mu_0(tt) = 0$, while $\mu_1(tt) = 1$ and $\mu_1(ht) = \mu_1(th) = \mu_1(hh) = 0$. Let $X_1$ and $X_2$ be the random variables representing the outcome of the first and second coin tosses, respectively. Clearly there is a functional dependence between $X_1$ and $X_2$, but it is easy to check that $X_1$ and $X_2$ are type-1 independent with respect to $\mathcal{P}$. Moreover, noninteraction holds: $NI_{\text{Pl}}(X_1 = i, X_2 = j)$ holds for $i, j \in \{h, t\}$. On the other hand, $I_{\text{Pl}_{\mathcal{P}}}(X_1, X_2)$ does not hold. For example, $f_{X_1 = h}(1) = 0$ while $f_{X_1 = h | X_2 = h}(1) = *$.[9]

The difference between noninteraction (i.e., type-1 independence) and the definition of independence used in this paper in the context of sets of probability measures can be summarized as follows. $U$ and $V$ do not interact with respect to $\mathcal{P}$ if $U$ and $V$ are independent with respect to all measures $\mu \in \mathcal{P}$. On the other hand, $U$ and $V$ are independent with respect to $\mathcal{P}$ if (1) $U$ and $V$ are independent for all measures $\mu \in \mathcal{P}$ such that $\mu(U) > 0$ and $\mu(V) > 0$ and (2) $\mu(U) = 0$ iff $\mu(V) = 0$ for all $\mu \in \mathcal{P}$. The definition of independence used here is thus more restrictive; it does not ignore the measures that give $U$ or $V$ probability 0 when determining independence. The difference between the two approaches is illustrated in the example in the previous paragraph.

---

9. As Peter Walley [private communication, 2000] points out, this example is somewhat misleading. The definition of independence with respect to $\text{Pl}_{\mathcal{P}}$ produces the same counterintuitive behavior as type-1 independence if the probabilities are modified slightly so as to make them positive, i.e., when there is "almost functional dependence" between the two variables. For example, suppose that the coin in the example is known to either land heads with probability .99 or .01 (rather than 1 and 0, as in the example). Let $\mu_0'$ and $\mu_1'$ be the obvious modifications of $\mu_0$ and $\mu_1$ required to represent this situation, and let $\mathcal{P}' = \{\mu_0', \mu_1'\}$. It is easy to check that $X_1$ and $X_2$ continue to be type-1 independent, and noninteraction continues to hold, but now $I_{\text{Pl}_{\mathcal{P}'}}(X_1, X_2)$ also holds. The real problem is that this representation of uncertainty does not enable learning.

As the variant of the example considered in Footnote 9 shows though, neither definition can completely claim to represent the intuition that if $U$ is independent of $V$, then learning $U$ gives no information about $V$. If the coin in the example is known to land heads with probability either .99 or .01, then seeing the first coin toss land heads certainly seems to give information about the second coin toss, even though both definitions would declare the events independent. However, the definition of independence used here does have the advantage of leading to an algebraic cps, which means, as is shown in the next section, that using it leads to a representation of sets of probability measures that can be represented as a Bayesian network.

## 5. Bayesian Networks

Throughout this section, I assume that we start with a set $W$ of possible worlds characterized by a set $\mathcal{X} = \{X_1, \ldots, X_n\}$ of $n$ binary random variables. That is, a world in $W$ is a tuple $(x_1, \ldots, x_n)$ with $x_i \in \{0, 1\}$, and $X_i(x_1, \ldots, x_n) = x_i$; that is, the value of $X_i$ in world $w = (x_1, \ldots, x_n)$ is $w_i$.[10] The goal of this section is to show that many of the tools of Bayesian network technology can be applied in this setting. The proofs of the main results all proceed in essentially the same spirit as well-known results for probabilistic Bayesian networks (see (Geiger & Pearl, 1988; Geiger et al., 1990; Verma, 1986)).

### 5.1 Qualitative Bayesian Networks

As usual, a (qualitative) *Bayesian network* (over $\mathcal{X}$) is a *dag* whose nodes are labeled by variables in $\mathcal{X}$. The standard notion of a Bayesian network representing a probability measure (Pearl, 1988) can be generalized in the obvious way to plausibility.

**Definition 5.1:** Given a qualitative Bayesian network $G$, let $\mathrm{Par}_G(X)$ be the *parents* of the random variable $X$ in $G$; let $\mathrm{Des}_G(X)$ be all the *descendants* of $X$, that is, $X$ and all those nodes $Y$ such that $X$ is an ancestor of $Y$; let $\mathrm{ND}_G(X)$, the *nondescendants of $X$*, consist of $\mathcal{X} - \mathrm{Des}_G(X)$. Note that all ancestors of $X$ are nondescendants of $X$. The Bayesian network $G$ *is compatible with* the cps $(W, \mathcal{F}, \mathcal{F}', \mathrm{Pl})$ (or just *compatible with* $\mathrm{Pl}$, if the other components of the cps are clear from context) if $I_{\mathrm{Pl}}^{rv}(X, \mathrm{ND}_G(X)|\mathrm{Par}(X))$, that is, if $X$ is conditionally independent of its nondescendants given its parents, for all $X \in \mathcal{X}$. ∎

There is a standard way of constructing a Bayesian network that represents a probability measure (Pearl, 1988). I briefly review the construction here, since it works without change for an algebraic cpm. Given an algebraic cpm $\mathrm{Pl}$, let $Y_1, \ldots, Y_n$ be a permutation of the random variables in $\mathcal{X}$. Construct a qualitative Bayesian network $G_{\mathrm{Pl}, \langle Y_1, \ldots, Y_n \rangle}$ as follows: For each $k$, find a minimal subset of $\{Y_1, \ldots, Y_{k-1}\}$, call it $\mathbf{P}_k$, such that $I_{\text{Pl}}^{rv}(\{Y_1, \ldots, Y_{k-1}\}, Y_k | \mathbf{P}_k)$. Then add edges from each of the nodes in $\mathbf{P}_k$ to $Y_k$. Verma (1986) shows that this construction gives a Bayesian network that is compatible with Pl in the case that Pl is a probability measure; his proof depends only on CIRV1–4. Thus, the construction works for algebraic cpms.

---

10. The assumption that the random variables are binary is just for ease of exposition. It is easy to generalize the results to the case where $\mathcal{R}(X_i)$ is finite for each $X_i$; there is no need to assume that $\mathcal{R}(X_i)$ is a subset of the reals.

**Theorem 5.2:** $G_{\text{Pl}, \langle Y_1, \ldots, Y_n \rangle}$ *is compatible with Pl.*

**Proof:** For ease of notation in the proof, I write $G$ instead of $G_{\text{Pl}, \langle Y_1, \ldots, Y_n \rangle}$. Note that $Y_1, \ldots, Y_n$ represents a topological sort of $G$; edges always go from nodes in $\{Y_1, \ldots, Y_{k-1}\}$ to $Y_k$. It follows that $G$ is acyclic; i.e., it is a dag. The construction guarantees that $\mathbf{P}_k = \text{Par}_G(Y_k)$ and that $I_{\text{Pl}}^{rv}(\{Y_1, \ldots, Y_{k-1}\}, Y_k | \text{Par}_G(Y_k))$. It follows from results of (Verma, 1986) (and is not hard to verify directly) that $I_{\text{Pl}}^{rv}(\text{ND}_G(Y_k), Y_k | \text{Par}_G(Y_k))$ can be proved using only CIRV1–4. The result now follows from Theorem 4.5. ∎

## 5.2 Quantitative Bayesian Networks

A qualitative Bayesian network $G$ gives qualitative information about dependence and independence, but does not actually give the values of the conditional plausibilities. To provide the more quantitative information, we associate with each node $X$ in $G$ a *conditional plausibility table (cpt)* that quantifies the effects of the parents of $X$ on $X$. A cpt for $X$ gives, for each setting of $X$'s parents in $G$, the plausibility that $X = 0$ and $X = 1$ given that setting. For example, if $X$'s parents in $G$ are $Y$ and $Z$, then the cpt for $X$ would have an entry denoted $d_{X=i|Y=j \cap Z=k}$ for all $(i, j, k) \in \{0, 1\}^3$. As the notation is meant to suggest, $d_{X=i|Y=j \cap Z=k} = \text{Pl}(X = i | Y = j \cap Z = k)$ for the plausibility measure Pl represented by $G$.[11] For each fixed $j$ and $k$, we assume that $x_{0jk} \oplus x_{1jk} = \top$. A *quantitative Bayesian network* is a pair $(G, f)$ consisting of a qualitative Bayesian network $G$ and a function $f$ that associates with each node $X$ in $G$ a cpt for $X$.

**Definition 5.3:** A quantitative Bayesian network $(G, f)$ *represents* Pl if $G$ is compatible with Pl and the cpts agree with Pl, in the sense that, for each random variable $X$, the entry $d_{X=i|Y_1=j_1, \ldots, Y_k=j_k}$ in the cpt is $\text{Pl}(X = i | Y_1 = j_1 \cap \ldots \cap Y_k = j_k)$ if $Y_1 = j_1 \cap \ldots \cap Y_k = j_k \in \mathcal{F}'$. (It does not matter what $d_{X=i|Y_1=j_1, \ldots, Y_k=j_k}$ is if $Y_1 = j_1 \cap \ldots \cap Y_k = j_k \notin \mathcal{F}'$.) ∎

Given a cpm Pl, it is easy to construct a quantitative Bayesian network $(G, f)$ that represents Pl: simply construct $G$ that is compatible with Pl as in Theorem 5.2 and define $f$ appropriately, using Pl. The more interesting question is whether there is a unique algebraic cpm determined by a quantitative Bayesian network. As stated, this question is somewhat undetermined. The numbers in a quantitative network do not say what $\oplus$ and $\otimes$ ought to be for the algebraic cpm.

A reasonable way to make the question more interesting is the following. Recall that, for the purposes of this section, I have taken $W$ to consist of the $2^n$ worlds characterized by the $n$ binary random variables in $\mathcal{X}$. Let $\mathcal{PL}_{D, \oplus, \otimes}$ consist of all standard cps's of the form $(W, \mathcal{F}, \mathcal{F}', \text{Pl})$, where $\mathcal{F} = 2^W$, so that all subsets of $W$ are measurable, the range of Pl is $D$, and Pl is algebraic with respect to $\oplus$ and $\otimes$. Thus, for example, $\mathcal{PL}_{IN^*,\min,+}$ consists of all conditional ranking functions on $W$ defined from unconditional ranking functions by the construction in Section 2. Since a cps $(W, \mathcal{F}, \mathcal{F}', \mathrm{Pl}) \in \mathcal{PL}_{D,\oplus,\otimes}$ is determined by Pl, I often abuse notation and write $\mathrm{Pl} \in \mathcal{PL}_{D,\oplus,\otimes}$.

---

11. Of course, if the random variables are not binary, $i, j, k$ have to range over all possible values for the random variables.

With this notation, the question becomes whether a quantitative Bayesian network $(G, f)$ such that the entries in the cpts are in $D$ determines a unique element in $\mathcal{PL}_{D,\oplus,\otimes}$. As I now show, the answer is yes, provided $(D, \oplus, \otimes)$ satisfies some conditions. Characterizing the conditions on $(D, \oplus, \otimes)$ required for this result turns out to be a little subtle. Indeed, it is somewhat surprising how many assumptions are required to reproduce the simple arguments that are required in the case of probability.

**Definition 5.4:** $(D, \oplus, \otimes)$ is a *BN-compatible domain (with respect to $\mathcal{PL}_{D,\oplus,\otimes}$)* if there are sets $D(\otimes) \subseteq D \times D$ and $D(\oplus) \subseteq D \cup D^2 \cup D^3 \cup \ldots$ satisfying the following properties:

BN1. $\oplus$ and $\otimes$ are commutative and associative.

BN2. For all $d \in D$, $(\top, d), (\bot, d) \in D(\otimes)$, $(\bot, d) \in D(\oplus)$, $\top \otimes d = d$, $\bot \otimes d = \bot$, and $\bot \oplus d = d$.

BN3. $\otimes$ distributes over $\oplus$; more precisely, $a \otimes (b_1 \oplus \cdots \oplus b_n) = (a \otimes b_1) \oplus \cdots \oplus (a \otimes b_n)$ if $(a, b_1), \ldots, (a, b_n), (a, b_1 \oplus \cdots \oplus b_n) \in D(\otimes)$ and $(b_1, \ldots, b_n), (a \otimes b_1, \ldots, a \otimes b_n) \in D(\oplus)$; moreover, $(a_1 \oplus \cdots \oplus a_n) \otimes b = a_1 \otimes b \oplus \cdots \oplus a_n \otimes b$ if $(a_1, \ldots, a_n), (a_1 \otimes b, \ldots, a_n \otimes b) \in \mathcal{D}(\oplus)$ and $(a_1 \oplus \cdots \oplus a_n, b), (a_1, b), \ldots, (a_n, b) \in D(\otimes)$.

BN4. If $(a, c), (b, c) \in D(\otimes)$, $a \otimes c \leq b \otimes c$, and $c \neq \bot$, then $a \leq b$.

BN5. If $(d_1, \ldots, d_k) \in D(\oplus)$ and $d_1 \oplus \cdots \oplus d_k \leq d$, then there exists $(d_1', \ldots, d_k') \in D(\oplus)$ such that $(d_1', d), \ldots, (d_k', d), (d_1' \oplus \cdots \oplus d_k', d) \in D(\otimes)$, $d_i = d_i' \otimes d$, for $i = 1, \ldots, k$, and $d_1 \oplus \cdots \oplus d_k = (d_1' \oplus \cdots \oplus d_k') \otimes d$.

BN6. $D(\oplus)$ is closed under permutations and prefixes, so that if $(x_1, \ldots, x_k) \in D(\oplus)$ and $\pi$ is a permutation of $(1, \ldots, k)$, then $(x_{\pi(1)}, \ldots, x_{\pi(k)}) \in D(\oplus)$ and if $k' \leq k$, then $(x_1, \ldots, x_{k'}) \in D(\oplus)$; moreover $D(\oplus) \supseteq D$.

BN7. If $(d_1, \ldots, d_k), (d_1', \ldots, d_m') \in D(\oplus)$, $(d_i, d_j') \in D(\otimes)$ for $i = 1, \ldots, k$, $j = 1, \ldots, m$, then $(d_1 \otimes d_1', \ldots, d_1 \otimes d_m', \ldots, d_k \otimes d_1', \ldots, d_k \otimes d_m') \in D(\oplus)$.

BN8. If $(d_1, \ldots, d_k) \in D(\oplus)$ and $k' \leq k$, then $d_1 \oplus \cdots \oplus d_{k'} \leq d_1 \oplus \cdots \oplus d_k$.

Note that all the representations of uncertainty we have considered so far have associated with them BN-compatible domains. Indeed, the definitions of $D(\oplus)$, $D(\otimes)$, $\oplus$, and $\otimes$ in each case are given in the proof of Proposition 3.2. For example, for $\mathcal{PL}_{[0,1],\max,\min}$, the set of conditional possibility measures determined by unconditional possibility measures, $D(\oplus) = [0, 1] \times [0, 1]$, while $D(\otimes)$ consists of all pairs $(a, b) \in [0, 1] \times [0, 1]$ such that $a < b$ or $a = 1$. I leave it to the reader to check that, in all these cases, BN1–8 hold.

Given a tuple $\mathbf{x} = (x_1, \ldots, x_n) \in [0, 1]^n$, let $d_{X_i, G, \mathbf{x}}$ denote the value $d_{X_i = x_i \mid \mathrm{Par}_G(X_i) = \mathbf{y}}$, where $\mathbf{y}$ is the restriction of $\mathbf{x}$ to the variables in $\mathrm{Par}_G(X_i)$.

**Definition 5.5:** If $(D, \oplus, \otimes)$ is BN-compatible, then a quantitative Bayesian network $(G, f)$ is $(D, \oplus, \otimes)$-*representable* if the values of the cpts for $G$ lie in $D$ and the following properties hold:

R1. For every node $X$ in $G$ and every setting $\mathbf{y}$ of $\mathrm{Par}_G(X)$, $(d_{X=0|\mathrm{Par}_G(X)=\mathbf{y}}, d_{X=1|\mathrm{Par}_G(X)=\mathbf{y}}) \in Dom(\oplus)$ and

$$d_{X=0|\mathrm{Par}_G(X)=\mathbf{y}} \oplus d_{X=1|\mathrm{Par}_G(X)=\mathbf{y}} = \top.$$

R2. Suppose $Y_1, \ldots, Y_n$ is a topological sort of the nodes in $G$. Then for all $\mathbf{y} \in \{0,1\}^n$ and all $1 \leq j < k \leq n$, $(d_{Y_j, G, \mathbf{y}}, d_{Y_{j+1}, G, \mathbf{y}} \otimes \cdots \otimes d_{Y_k, G, \mathbf{y}}) \in D(\otimes)$ and $(d_{Y_j, G, \mathbf{y}} \otimes \cdots \otimes d_{Y_{k-1}, G, \mathbf{y}}, d_{Y_k, G, \mathbf{y}}) \in D(\otimes)$.

R1 is the obvious analogue of the requirement in the probabilistic case that the entries of the cpt for $X$, for a fixed setting of $X$'s parents, add up to 1. R2 essentially says that certain terms (the ones required to compute the plausibility of $\mathbf{Y} = \mathbf{y}$ for $\mathbf{Y} = \langle Y_1, \ldots, Y_n \rangle$) are required to be in $D(\otimes)$, so that it makes sense to take their product. Since $D(\otimes) = [0,1] \times [0,1]$ in the case of probability, there is no need to make this requirement explicit. However, it is necessary for other representations of uncertainty.

The following result shows that, as the name suggests, there is a unique cpm that represents a representable quantitative Bayesian network.

**Theorem 5.6:** *If $(G, f)$ is $(D, \oplus, \otimes)$-representable, then there is a unique cpm $Pl \in \mathcal{PL}_{D, \oplus, \otimes}$ such that $(G, f)$ represents $Pl$.*

### 5.3 D-Separation

Just as in the case of probability, conditional independencies can be read off the Bayesian network using the criterion of d-separation (Pearl, 1988). Recall that a set $\mathbf{X}$ of nodes in $G = (V, E)$, is *d-separated* from a set $\mathbf{Y}$ of nodes by a set $\mathbf{Z}$ of nodes in $G$, written $d\text{-}sep_G(\mathbf{X}, \mathbf{Y} | \mathbf{Z})$, if, for every $X \in \mathbf{X}$, $Y \in \mathbf{Y}$, and a *trail* from $X$ to $Y$ (that is, a sequence $(X_0, \ldots, X_k)$ of nodes in $G$ such that $X_0 = X$, $X_k = Y$ and either $(X_i, X_{i+1})$ or $(X_{i+1}, X_i)$ is a directed edge in $G$) and a node $X_i$ on the trail with $0 < i < k$ such that either:

(a) $X_i \in \mathbf{Z}$ and there is an arrow leading into $X_i$ and an arrow leading out (i.e., either $(X_{i-1}, X_i), (X_i, X_{i+1}) \in E$ or $(X_i, X_{i-1}), (X_{i+1}, X_i) \in E$

(b) $X_i \in \mathbf{Z}$ and $X_i$ is a tail-to-tail node (i.e., $(X_i, X_{i-1}), (X_i, X_{i+1}) \in E$)

(c) $X_i$ is a head to head node (i.e., $(X_{i-1}, X_i), (X_{i+1}, X_i) \in E$), and neither $X_i$ nor any of its descendants are in $\mathbf{Z}$.

Let $\Sigma_{G, \mathrm{Pl}}$ consist of all statements of the form $I_{\mathrm{Pl}}^{rv}(X, \mathrm{ND}_G(X) | \mathrm{Par}_G(X))$. Let $\mathcal{PL}_{D, \oplus, \otimes}$ be an arbitrary collection of cps's of the form $(W, \mathcal{F}, \mathcal{F}', \mathrm{Pl})$ where all components other than $\mathrm{Pl}$ are fixed, and the plausibility measures $\mathrm{Pl}$ all have the same range $D$ of plausibility values. Consider the following three statements:

1. $d\text{-}sep_G(\mathbf{X}, \mathbf{Y} | \mathbf{Z})$.

2. $I_{\mathrm{Pl}}^{rv}(\mathbf{X}, \mathbf{Y} | \mathbf{Z})$ is provable from CIRV1–4 and $\Sigma_{G, \mathrm{Pl}}$.

3. $I_{\mathrm{Pl}}^{rv}(\mathbf{X}, \mathbf{Y}|\mathbf{Z})$ holds for every plausibility measure in $\mathcal{PL}_{D,\oplus,\otimes}$ compatible with $G$.

The implication from 1 to 2 is proved in (Geiger et al., 1990; Verma, 1986).

**Theorem 5.7:** (Geiger et al., 1990; Verma, 1986) *If d-sep$_G$($\mathbf{X}, \mathbf{Y}|\mathbf{Z}$), then $I_{\mathrm{Pl}}^{rv}(\mathbf{X}, \mathbf{Y}|\mathbf{Z})$ is provable from CIRV1–4 and $\Sigma_{G,\mathrm{Pl}}$.*

It is immediate from Theorem 4.5 that the implication from 2 to 3 holds for algebraic cpms.

**Corollary 5.8:** *If $I_{\mathrm{Pl}}^{rv}(\mathbf{X}, \mathbf{Y}|\mathbf{Z})$ is provable from CIRV1–4 and $\Sigma_{G,\mathrm{Pl}}$, then $I_{\mathrm{Pl}}^{rv}(\mathbf{X}, \mathbf{Y}|\mathbf{Z})$ holds for every algebraic cpm Pl compatible with $G$.*

Finally, the implication from 3 to 1 for probability measures is proved in (Geiger & Pearl, 1988; Geiger et al., 1990). Here I generalize the proof to algebraic plausibility measures. Notice that to prove the implication from 3 to 1, it suffices to show that if $X$ is not d-separated from $Y$ by $\mathbf{Z}$ in $G$, then there is a plausibility measure $\mathrm{Pl} \in \mathcal{PL}_{D,\oplus,\otimes}$ such that $I_{\mathrm{Pl}}^{rv}(X, Y|\mathbf{Z})$ does not hold. To guarantee that such a plausibility measure exists in $\mathcal{PL}_{D,\oplus,\otimes}$, we have to ensure that there are "enough" plausibility measures in $\mathcal{PL}_{D,\oplus,\otimes}$ in the following technical sense.

**Definition 5.9:** A BN-compatible domain $(D, \oplus, \otimes)$ is *rich* if there exist $d, d' \in D$ such that (1) $(d, d') \in D(\oplus)$, (2) $d \oplus d' = \top$ and (3) if $x = x_1 \otimes \ldots \otimes x_k$, where each $x_i$ is either $d$ or $d'$ and $k < n$, then $(d, x)$, $(x, d)$, $(d', x)$, and $(x, d')$ are all in $D(\otimes)$ (intuitively, $D(\otimes)$ contains all products involving $d$ and $d'$ of length at most $n$). ∎

All the domains for the cps's we have considered are easily seen to be rich.

**Theorem 5.10:** *Suppose that plausibility measures in $\mathcal{PL}_{D,\oplus,\otimes}$ take values in a rich BN-compatible domain. Then if $I_{\mathrm{Pl}}^{rv}(\mathbf{X}, \mathbf{Y}|\mathbf{Z})$ holds for every plausibility measure in $\mathcal{PL}_{D,\oplus,\otimes}$ compatible with $G$, then d-sep$_G$($\mathbf{X}, \mathbf{Y}|\mathbf{Z}$).*

I remark that independence and d-separation for various approaches to representing sets of probability measures using Bayesian networks are discussed by Cozman (2000b, 2000a). However, the technical details are quite different from the approach taken here.

## 6. Conclusion

I have considered a general notion of conditional plausibility that generalizes all other standard notions of conditioning in the literature, and examined various requirements that could be imposed on conditional plausibility. One set of requirements, those that lead to algebraic cps's, was shown to suffice for the construction of Bayesian networks. Further assuming that the range $D$ of the plausibility measure is a BN-compatible domain suffices for all the more quantitative properties of Bayesian networks to hold and for d-separation to characterize the independencies. It should also be clear that standard constructions like belief propagation in Bayesian networks (Pearl, 1988) can also be applied to algebraic cps's with ranges that are BN-compatible, since they typically use only basic properties of conditioning, addition, and multiplication, all of which hold in BN-compatible domains (using $\oplus$ and $\otimes$). In particular, these results apply to sets to probability measures, provided that they are appropriately represented as plausibility measures. The particular representation of sets of probability measures advocated in this paper was also shown to have a number of other attractive properties.

The results of this paper show that Alg1–4 are *sufficient* conditions for representing a measure of uncertainty that is acceptable in a Bayesian network. They may not be necessary. It would be interesting to see if other natural conditions also suffice. Similarly, I have focused only on *acceptable* cps's, that is, ones that satisfy Acc1–4. Acc3 and Acc4 are nontrivial conditions; it would be of interest to see to what extent they could be weakened while still being able to prove results in the spirit of this paper. I leave these questions to future research.

## Appendix A. Proofs

In this section I give the proofs of Theorems 4.5, 5.6, and 5.10. I repeat the statement of the results for the convenience of the reader.

**Lemma A.1:** *Suppose that* $(W, \mathcal{F}, \mathcal{F}', Pl)$ *is a cps,* $A_1, \ldots, A_n$ *is a partition of* $W$, $X, A_1, \ldots, A_n \in \mathcal{F}$, *and* $Y \in \mathcal{F}'$. *Then*

$$Pl(X|Y) = \oplus_{\{i: A_i \cap Y \in \mathcal{F}'\}} Pl(X|A_i \cap Y) \otimes Pl(A_i|Y).^{12}$$

**Proof:** Using an easy induction argument, it follows from Alg1 that

$$\text{Pl}(X|Y) = \oplus_{i=1}^n \text{Pl}(X \cap A_i|Y).$$

If $A_i \cap Y \notin \mathcal{F}'$, then it follows from Acc4 that $\text{Pl}(A_i|Y) = \perp$. Thus, by CPl3, $\text{Pl}(X \cap A_i|Y) = \perp$. Using Lemma 3.3, it follows that

$$\text{Pl}(X|Y) = \oplus_{\{i: A_i \cap Y \in \mathcal{F}'\}} \text{Pl}(X \cap A_i|Y).$$

If $A_i \cap Y \in \mathcal{F}'$, then it follows from Alg2 that $\text{Pl}(X \cap A_i|Y) = \text{Pl}(X|A_i \cap Y) \otimes \text{Pl}(A_i|Y)$. Thus,

$$\text{Pl}(X|Y) = \oplus_{\{i: A_i \cap Y \in \mathcal{F}'\}} \text{Pl}(X|A_i \cap Y) \otimes \text{Pl}(A_i|Y),$$

as desired. ∎

**Theorem 4.5:** *CIRV1–4 hold for all algebraic cps's.*

**Proof:** CIRV1 is immediate from the fact that independence is symmetric.

For CIRV2, suppose that $I_{\text{Pl}}^{rv}(\mathbf{X}, \mathbf{Y} \cup \mathbf{Y}'|\mathbf{Z})$. We must show $I_{\text{Pl}}^{rv}(\mathbf{X}, \mathbf{Y}|\mathbf{Z})$. That is, we must show that $I_{\text{Pl}}(\mathbf{X} = \mathbf{x}, \mathbf{Y} = \mathbf{y}|\mathbf{Z} = \mathbf{z})$, for all $\mathbf{x}$, $\mathbf{y}$, and $\mathbf{z}$. This requires showing two things.

---

12. Notice that if $A_i \cap Y \in \mathcal{F}'$, then $\text{Pl}(X|A_i \cap Y) \otimes \text{Pl}(A_i|Y) = \text{Pl}(X \cap A_i|Y)$ by Alg2. Thus, the terms arising on the right-hand side of the equation in Lemma A.1 are in $Dom(\oplus)$. This means that there is no need to put in parentheses; $\oplus$ is associative on terms in $Dom(\oplus)$.

2(a). If $\mathbf{X} = \mathbf{x} \cap \mathbf{Z} = \mathbf{z} \in \mathcal{F}'$, then

$$\mathrm{Pl}(\mathbf{Y} = \mathbf{y}|\mathbf{X} = \mathbf{x} \cap \mathbf{Z} = \mathbf{z}) = \mathrm{Pl}(\mathbf{Y} = \mathbf{y}|\mathbf{Z} = \mathbf{z}).$$

2(b). If $\mathbf{Y} = \mathbf{y} \cap \mathbf{Z} = \mathbf{z} \in \mathcal{F}'$, then

$$\mathrm{Pl}(\mathbf{X} = \mathbf{x}|\mathbf{Y} = \mathbf{y} \cap \mathbf{Z} = \mathbf{z}) = \mathrm{Pl}(\mathbf{X} = \mathbf{x}|\mathbf{Z} = \mathbf{z}).$$

For 2(a), suppose that $\mathrm{Pl}(\mathbf{X} = \mathbf{x} \cap \mathbf{Z} = \mathbf{z}) \in \mathcal{F}'$. From $I_{\mathrm{Pl}}(\mathbf{X}, \mathbf{Y} \cup \mathbf{Y}'|\mathbf{Z})$, it follows that $I_{\mathrm{Pl}}(\mathbf{X} = \mathbf{x}, \mathbf{Y} = \mathbf{y} \cap \mathbf{Y}' = \mathbf{y}'|\mathbf{Z} = \mathbf{z})$ for all $\mathbf{y}'$. Hence

$$\mathrm{Pl}(\mathbf{Y} = \mathbf{y} \cap \mathbf{Y}' = \mathbf{y}'|\mathbf{X} = \mathbf{x} \cap \mathbf{Z} = \mathbf{z}) = \mathrm{Pl}(\mathbf{Y} = \mathbf{y} \cap \mathbf{Y}' = \mathbf{y}'|\mathbf{Z} = \mathbf{z}) \qquad (4)$$

for all $\mathbf{y}' \in \mathcal{R}(\mathbf{Y}')$. From (4) it follows that

$$\oplus_{\mathbf{y}'}\mathrm{Pl}(\mathbf{Y} = \mathbf{y} \cap \mathbf{Y}' = \mathbf{y}'|\mathbf{X} = \mathbf{x} \cap \mathbf{Z} = \mathbf{z}) = \oplus_{\mathbf{y}'}\mathrm{Pl}(\mathbf{Y} = \mathbf{y} \cap \mathbf{Y}' = \mathbf{y}'|\mathbf{Z} = \mathbf{z}).$$

Thus,

$$\mathrm{Pl}(\cup_{\mathbf{y}'}\mathbf{Y} = \mathbf{y} \cap \mathbf{Y}' = \mathbf{y}'|\mathbf{X} = \mathbf{x} \cap \mathbf{Z} = \mathbf{z}) = \mathrm{Pl}(\cup_{\mathbf{y}'}\mathbf{Y} = \mathbf{y} \cap \mathbf{Y}' = \mathbf{y}'|\mathbf{Z} = \mathbf{z}).$$

Since $\cup_{\mathbf{y}'}(\mathbf{Y} = \mathbf{y} \cap \mathbf{Y}' = \mathbf{y}') = \mathbf{Y} = \mathbf{y}$, 2(a) holds.

For 2(b), from $I_{\mathrm{Pl}}^{vp}(\mathbf{X}, \mathbf{Y} \cup \mathbf{Y}'|\mathbf{Z})$, it follows that if $\mathbf{Y} = \mathbf{y} \cap \mathbf{Y}' = \mathbf{y}' \cap \mathbf{Z} = \mathbf{z} \in \mathcal{F}'$, then

$$\mathrm{Pl}(\mathbf{X} = \mathbf{x}|\mathbf{Y} = \mathbf{y} \cap \mathbf{Y}' = \mathbf{y}' \cap \mathbf{Z} = \mathbf{z}) = \mathrm{Pl}(\mathbf{X} = \mathbf{x}|\mathbf{Z} = \mathbf{z}). \qquad (5)$$

From (5) and Lemma A.1, it follows that

$$\begin{aligned}
&\mathrm{Pl}(\mathbf{X} = \mathbf{x}|\mathbf{Y} = \mathbf{y} \cap \mathbf{Z} = \mathbf{z}) \\
=\ &\oplus_{\{\mathbf{y}':\mathbf{Y}=\mathbf{y}\cap\mathbf{Y}'=\mathbf{y}'\cap\mathbf{Z}=\mathbf{z}\in\mathcal{F}'\}}\mathrm{Pl}(\mathbf{X} = \mathbf{x}|\mathbf{Y} = \mathbf{y} \cap \mathbf{Y}' = \mathbf{y}' \cap \mathbf{Z} = \mathbf{z}) \otimes \mathrm{Pl}(\mathbf{Y}' = \mathbf{y}'|\mathbf{Y} = \mathbf{y} \cap \mathbf{Z} = \mathbf{z}) \\
=\ &\oplus_{\{\mathbf{y}':\mathbf{Y}=\mathbf{y}\cap\mathbf{Y}'=\mathbf{y}'\cap\mathbf{Z}=\mathbf{z}\in\mathcal{F}'\}}\mathrm{Pl}(\mathbf{X} = \mathbf{x}|\mathbf{Z} = \mathbf{z}) \otimes \mathrm{Pl}(\mathbf{Y}' = \mathbf{y}'|\mathbf{Y} = \mathbf{y} \cap \mathbf{Z} = \mathbf{z}).
\end{aligned} \qquad (6)$$

By Acc4, it follows that if $\mathbf{Y} = \mathbf{y} \cap \mathbf{Y}' = \mathbf{y}' \cap \mathbf{Z} = \mathbf{z} \notin \mathcal{F}'$, then $\mathrm{Pl}(\mathbf{Y}' = \mathbf{y}'|\mathbf{Y} = \mathbf{y} \cap \mathbf{Z} = \mathbf{z}) = \bot$. Thus, by Lemma 3.3, Alg1, CPl2, and CPl4,

$$\begin{aligned}
&\oplus_{\{\mathbf{y}':\mathbf{Y}=\mathbf{y}\cap\mathbf{Y}'=\mathbf{y}'\cap\mathbf{Z}=\mathbf{z}\in\mathcal{F}'\}}\mathrm{Pl}(\mathbf{Y}' = \mathbf{y}'|\mathbf{Y} = \mathbf{y} \cap \mathbf{Z} = \mathbf{z}) \\
=\ &\oplus_{\mathbf{y}'}\mathrm{Pl}(\mathbf{Y}' = \mathbf{y}'|\mathbf{Y} = \mathbf{y} \cap \mathbf{Z} = \mathbf{z}) \\
=\ &\mathrm{Pl}(W|\mathbf{Y} = \mathbf{y} \cap \mathbf{Z} = \mathbf{z}) \\
=\ &\top.
\end{aligned} \qquad (7)$$

The next step is to apply distributivity (Alg3) to the last line of (6). To do this, we must show that certain tuples are in $Dom(\oplus)$ and $Dom(\otimes)$, respectively. Since

$$(\mathrm{Pl}(\mathbf{X} = \mathbf{x}|\mathbf{Y} = \mathbf{y} \cap \mathbf{Y}' = \mathbf{y}' \cap \mathbf{Z} = \mathbf{z}), \mathrm{Pl}(\mathbf{Y}' = \mathbf{y}'|\mathbf{Y} = \mathbf{y} \cap \mathbf{Z} = \mathbf{z}) \in Dom(\otimes),$$

from (5) it follows that

$$(\mathrm{Pl}(\mathbf{X} = \mathbf{x}|\mathbf{Z} = \mathbf{z}), \mathrm{Pl}(\mathbf{Y}' = \mathbf{y}'|\mathbf{Y} = \mathbf{y} \cap \mathbf{Z} = \mathbf{z})) \in Dom(\otimes).$$

If $\{\mathbf{y}'_{i_1}, \ldots, \mathbf{y}'_{i_k}\} = \{\mathbf{y}' \in \mathcal{R}(\mathbf{Y}') : \mathbf{Y} = \mathbf{y} \cap \mathbf{Y}' = \mathbf{y}' \cap \mathbf{Z} = \mathbf{z} \in \mathcal{F}'\}$, then clearly

$$(\mathrm{Pl}(\mathbf{Y}' = \mathbf{y}'_{i_1} | \mathbf{Y} = \mathbf{y} \cap \mathbf{Z} = \mathbf{z}), \ldots, \mathrm{Pl}(\mathbf{Y}' = \mathbf{y}'_{i_k} | \mathbf{Y} = \mathbf{y} \cap \mathbf{Z} = \mathbf{z}) \in Dom(\oplus).$$

Moreover, using (5) again and Alg2, it follows that

$$\mathrm{Pl}(\mathbf{X} = \mathbf{x} | \mathbf{Z} = \mathbf{z}) \otimes \mathrm{Pl}(\mathbf{Y}' = \mathbf{y}'_{i_k} | \mathbf{Y} = \mathbf{y} \cap \mathbf{Z} = \mathbf{z}) = \mathrm{Pl}(\mathbf{X} = \mathbf{x} \cap \mathbf{Y}' = \mathbf{y}'_{i_k} | \mathbf{Y} = \mathbf{y} \cap \mathbf{Z} = \mathbf{z}).$$

Thus, $(\mathrm{Pl}(\mathbf{X} = \mathbf{x} | \mathbf{Z} = \mathbf{z}) \otimes \mathrm{Pl}(\mathbf{Y}' = \mathbf{y}'_{i_1} | \mathbf{Y} = \mathbf{y} \cap \mathbf{Z} = \mathbf{z}), \ldots, \mathrm{Pl}(\mathbf{X} = \mathbf{x} | \mathbf{Z} = \mathbf{z}) \otimes \mathrm{Pl}(\mathbf{Y}' = \mathbf{y}'_{i_k} | \mathbf{Y} = \mathbf{y} \cap \mathbf{Z} = \mathbf{z}) \in Dom(\oplus)$. Finally, since (7) shows that $\oplus_{\{\mathbf{y}' : \mathbf{Y} = \mathbf{y} \cap \mathbf{Y}' = \mathbf{y}' \cap \mathbf{Z} = \mathbf{z} \in \mathcal{F}'\}} = \top$ and, by the proof of Lemma 3.4, $(d, \top) \in Dom(\otimes)$ for all $d \in Range(\mathrm{Pl})$, it follows that

$$(\mathrm{Pl}(\mathbf{X} | \mathbf{Z} = \mathbf{z}), \oplus_{\{\mathbf{y}' : \mathbf{Y} = \mathbf{y} \cap \mathbf{Y}' = \mathbf{y}' \cap \mathbf{Z} = \mathbf{z} \in \mathcal{F}'\}} \mathrm{Pl}(\mathbf{Y}' = \mathbf{y}' | \mathbf{Y} = \mathbf{y} \cap \mathbf{Z} = \mathbf{z})) \in Dom(\otimes).$$

It now follows, using Alg3, (7), and Lemma 3.4, that

$$
\begin{aligned}
& \oplus_{\{\mathbf{y}' : \mathbf{Y} = \mathbf{y} \cap \mathbf{Y}' = \mathbf{y}' \cap \mathbf{Z} = \mathbf{z} \in \mathcal{F}'\}} \mathrm{Pl}(\mathbf{X} = \mathbf{x} | \mathbf{Z} = \mathbf{z}) \otimes \mathrm{Pl}(\mathbf{Y}' = \mathbf{y}' | \mathbf{Y} = \mathbf{y} \cap \mathbf{Z} = \mathbf{z}) \\
= \ & \mathrm{Pl}(\mathbf{X} = \mathbf{x} | \mathbf{Z} = \mathbf{z}) \otimes (\oplus_{\{\mathbf{y}' : \mathbf{Y} = \mathbf{y} \cap \mathbf{Y}' = \mathbf{y}' \cap \mathbf{Z} = \mathbf{z} \in \mathcal{F}'\}} \mathrm{Pl}(\mathbf{Y}' = \mathbf{y}' | \mathbf{Y} = \mathbf{y} \cap \mathbf{Z} = \mathbf{z})) \\
= \ & \mathrm{Pl}(\mathbf{X} = \mathbf{x} | \mathbf{Z} = \mathbf{z}) \otimes \top \\
= \ & \mathrm{Pl}(\mathbf{X} = \mathbf{x} | \mathbf{Z} = \mathbf{z}).
\end{aligned}
$$

Thus, from (6), it follows that $\mathrm{Pl}(\mathbf{X} = \mathbf{x} | \mathbf{Y} = \mathbf{y} \cap \mathbf{Z} = \mathbf{z}) = \mathrm{Pl}(\mathbf{X} = \mathbf{x} | \mathbf{Z} = \mathbf{z})$. This completes the proof of 2(b) and CIRV2.

For CIRV3, suppose that $I^{rv}_{\mathrm{Pl}}(\mathbf{X}, \mathbf{Y} \cup \mathbf{Y}' | \mathbf{Z})$. We must show that $I^{rv}_{\mathrm{Pl}}(\mathbf{X}, \mathbf{Y} | \mathbf{Y}' \cup \mathbf{Z})$. This again requires showing two things:

3(a). If $\mathbf{X} = \mathbf{x} \cap \mathbf{Y}' = \mathbf{y}' \cap \mathbf{Z} = \mathbf{z} \in \mathcal{F}'$, then

$$\mathrm{Pl}(\mathbf{Y} = \mathbf{y} | \mathbf{X} = \mathbf{x} \cap \mathbf{Y}' = \mathbf{y}' \cap \mathbf{Z} = \mathbf{z}) = \mathrm{Pl}(\mathbf{Y} = \mathbf{y} | \mathbf{Y}' = \mathbf{y}' \cap \mathbf{Z} = \mathbf{z}).$$

3(b). If $\mathbf{Y} = \mathbf{y} \cap \mathbf{Y}' = \mathbf{y}' \cap \mathbf{Z} = \mathbf{z} \in \mathcal{F}'$, then

$$\mathrm{Pl}(\mathbf{X} = \mathbf{x} | \mathbf{Y} = \mathbf{y} \cap \mathbf{Y}' = \mathbf{y}' \cap \mathbf{Z} = \mathbf{z}) = \mathrm{Pl}(\mathbf{X} = \mathbf{x} | \mathbf{Y}' = \mathbf{y}' \cap \mathbf{Z} = \mathbf{z}).$$

For 3(a), suppose that $\mathbf{X} = \mathbf{x} \cap \mathbf{Y}' = \mathbf{y}' \cap \mathbf{Z} = \mathbf{z} \in \mathcal{F}'$. Thus, by Acc3, $\mathbf{X} = \mathbf{x} \cap \mathbf{Z} = \mathbf{z} \in \mathcal{F}'$. Since $I^{rv}_{\mathrm{Pl}}(\mathbf{X}, \mathbf{Y} \cup \mathbf{Y}' | \mathbf{Z})$, it follows that

$$\mathrm{Pl}(\mathbf{Y} = \mathbf{y}'' \cap \mathbf{Y}' = \mathbf{y}' | \mathbf{X} = \mathbf{x} \cap \mathbf{Z} = \mathbf{z}) = \mathrm{Pl}(\mathbf{Y} = \mathbf{y}'' \cap \mathbf{Y}' = \mathbf{y}' | \mathbf{Z} = \mathbf{z}) \qquad (8)$$

for all $\mathbf{y}'' \in \mathcal{R}(\mathbf{Y})$. Applying Alg2 to each side of (8), it follows that

$$
\begin{aligned}
& \mathrm{Pl}(\mathbf{Y} = \mathbf{y} | \mathbf{Y}' = \mathbf{y}' \cap \mathbf{X} = \mathbf{x} \cap \mathbf{Z} = \mathbf{z}) \otimes \mathrm{Pl}(\mathbf{Y}' = \mathbf{y}' | \mathbf{X} = \mathbf{x} \cap \mathbf{Z} = \mathbf{z}) \\
= \ & \mathrm{Pl}(\mathbf{Y} = \mathbf{y} | \mathbf{Y}' = \mathbf{y}' \cap \mathbf{Z} = \mathbf{z}) \otimes \mathrm{Pl}(\mathbf{Y}' = \mathbf{y}' | \mathbf{Z} = \mathbf{z}).
\end{aligned}
$$

Thus, to prove 3(a), it follows from Alg4 that it suffices to show that

$$\mathrm{Pl}(\mathbf{Y}' = \mathbf{y}' | \mathbf{X} = \mathbf{x} \cap \mathbf{Z} = \mathbf{z}) = \mathrm{Pl}(\mathbf{Y}' = \mathbf{y}' | \mathbf{Z} = \mathbf{z}) \neq \perp.$$

But by (8) and Alg1, it follows that

$$
\begin{aligned}
& \mathrm{Pl}(\mathbf{Y}' = \mathbf{y}' | \mathbf{X} = \mathbf{x} \cap \mathbf{Z} = \mathbf{z}) \\
= \; & \oplus_{\mathbf{y}'' \in \mathcal{R}(\mathbf{Y})} \mathrm{Pl}(\mathbf{Y} = \mathbf{y}'' \cap \mathbf{Y}' = \mathbf{y}' | \mathbf{X} = \mathbf{x} \cap \mathbf{Z} = \mathbf{z}) \\
= \; & \oplus_{\mathbf{y}'' \in \mathcal{R}(\mathbf{Y})} \mathrm{Pl}(\mathbf{Y} = \mathbf{y}'' \cap \mathbf{Y}' = \mathbf{y}' | \mathbf{Z} = \mathbf{z}) \\
= \; & \mathrm{Pl}(\mathbf{Y}' = \mathbf{y}' | \mathbf{Z} = \mathbf{z}),
\end{aligned}
$$

as desired. Moreover, since $\mathbf{X} = \mathbf{x} \cap \mathbf{Y}' = \mathbf{y}' \cap \mathbf{Z} = \mathbf{z} \in \mathcal{F}'$, it follows from Acc4 that $\mathrm{Pl}(\mathbf{Y}' = \mathbf{y}' | \mathbf{Z} = \mathbf{z}) \neq \bot$.

For 3(b), suppose that $\mathbf{Y} = \mathbf{y} \cap \mathbf{Y}' = \mathbf{y}' \cap \mathbf{Z} = \mathbf{z} \in \mathcal{F}'$. Since $I^{rv}_{\mathrm{Pl}}(\mathbf{X}, \mathbf{Y} \cup \mathbf{Y}' | \mathbf{Z})$, it follows that

$$
\mathrm{Pl}(\mathbf{X} = \mathbf{x} | \mathbf{Y} = \mathbf{y} \cap \mathbf{Y}' = \mathbf{y}' \cap \mathbf{Z} = \mathbf{z}) = \mathrm{Pl}(\mathbf{X} = \mathbf{x} | \mathbf{Z} = \mathbf{z}).
$$

Thus, to prove 3(b), it suffices to show that

$$
\mathrm{Pl}(\mathbf{X} = \mathbf{x} | \mathbf{Y}' = \mathbf{y}' \cap \mathbf{Z} = \mathbf{z}) = \mathrm{Pl}(\mathbf{X} = \mathbf{x} | \mathbf{Z} = \mathbf{z}). \tag{9}
$$

Recall that we are assuming that $I^{rv}_{\mathrm{Pl}}(\mathbf{X}, \mathbf{Y} \cup \mathbf{Y}' | \mathbf{Z})$. By CIRV2, it follows that $I^{rv}_{\mathrm{Pl}}(\mathbf{X}, \mathbf{Y}' | \mathbf{Z})$. Thus, (9) is immediate from 2(b) (since $\mathbf{Y} = \mathbf{y} \cap \mathbf{Y}' = \mathbf{y}' \cap \mathbf{Z} = \mathbf{z} \in \mathcal{F}'$ implies that $\mathbf{Y}' = \mathbf{y}' \cap \mathbf{Z} = \mathbf{z} \in \mathcal{F}'$).

Finally, consider CIRV4. Suppose that $I^{rv}_{\mathrm{Pl}}(\mathbf{X}, \mathbf{Y} | \mathbf{Z})$ and $I^{rv}_{\mathrm{Pl}}(\mathbf{X}, \mathbf{Y}' | \mathbf{Y} \cup \mathbf{Z})$. We must show that $I^{rv}_{\mathrm{Pl}}(\mathbf{X}, \mathbf{Y} \cup \mathbf{Y}' | \mathbf{Z})$. As usual, this requires showing two things:

4(a). If $\mathbf{Y} = \mathbf{y} \cap \mathbf{Y}' = \mathbf{y}' \cap \mathbf{Z} = \mathbf{z} \in \mathcal{F}'$, then

$$
\mathrm{Pl}(\mathbf{X} = \mathbf{x} | \mathbf{Y} = \mathbf{y} \cap \mathbf{Y}' = \mathbf{y}' \cap \mathbf{Z} = \mathbf{z}) = \mathrm{Pl}(\mathbf{X} = \mathbf{x} | \mathbf{Z} = \mathbf{z}).
$$

4(b). If $\mathbf{X} = \mathbf{x} \cap \mathbf{Z} = \mathbf{z} \in \mathcal{F}'$, then

$$
\mathrm{Pl}(\mathbf{Y} = \mathbf{y} \cap \mathbf{Y}' = \mathbf{y}' | \mathbf{X} = \mathbf{x} \cap \mathbf{Z} = \mathbf{z}) = \mathrm{Pl}(\mathbf{Y} = \mathbf{y} \cap \mathbf{Y}' = \mathbf{y}' | \mathbf{Z} = \mathbf{z}).
$$

Both 4(a) and 4(b) are straightforward. For 4(a), suppose that $\mathbf{Y} = \mathbf{y} \cap \mathbf{Y}' = \mathbf{y}' \cap \mathbf{Z} = \mathbf{z} \in \mathcal{F}'$. Since $I^{rv}_{\mathrm{Pl}}(\mathbf{X}, \mathbf{Y}' | \mathbf{Y} \cup \mathbf{Z})$, it follows that

$$
\mathrm{Pl}(\mathbf{X} = \mathbf{x} | \mathbf{Y} = \mathbf{y} \cap \mathbf{Y}' = \mathbf{y}' \cap \mathbf{Z} = \mathbf{z}) = \mathrm{Pl}(\mathbf{X} = \mathbf{x} | \mathbf{Y} = \mathbf{y} \cap \mathbf{Z} = \mathbf{z}).
$$

And since $I^{rv}_{\mathrm{Pl}}(\mathbf{X}, \mathbf{Y} | \mathbf{Z})$, it follows that

$$
\mathrm{Pl}(\mathbf{X} = \mathbf{x} | \mathbf{Y} = \mathbf{y} \cap \mathbf{Z} = \mathbf{z}) = \mathrm{Pl}(\mathbf{X} = \mathbf{x} | \mathbf{Z} = \mathbf{z}).
$$

Thus we have 4(a).

For 4(b), suppose that $\mathbf{X} = \mathbf{x} \cap \mathbf{Z} = \mathbf{z} \in \mathcal{F}'$. There are now two cases to consider. If $\mathrm{Pl}(\mathbf{Y} = \mathbf{y} | \mathbf{X} = \mathbf{x} \cap \mathbf{Z} = \mathbf{z}) \neq \bot$ then, by Acc4, $\mathbf{X} = \mathbf{x} \cap \mathbf{Y} = \mathbf{y} \cap \mathbf{Z} = \mathbf{z} \in \mathcal{F}'$. Moreover, by Alg2,

$$
\mathrm{Pl}(\mathbf{Y} = \mathbf{y} \cap \mathbf{Y}' = \mathbf{y}' | \mathbf{X} = \mathbf{x} \cap \mathbf{Z} = \mathbf{z}) = \mathrm{Pl}(\mathbf{Y}' = \mathbf{y}' | \mathbf{X} = \mathbf{x} \cap \mathbf{Y} = \mathbf{y} \cap \mathbf{Z} = \mathbf{z}) \otimes \mathrm{Pl}(\mathbf{Y} = \mathbf{y} | \mathbf{X} = \mathbf{x} \cap \mathbf{Z} = \mathbf{z}). \tag{10}
$$

Since $I^{rv}_{\mathrm{Pl}}(\mathbf{X}, \mathbf{Y}' | \mathbf{Y} \cup \mathbf{Z})$, it follows that

$$
\mathrm{Pl}(\mathbf{Y}' = \mathbf{y}' | \mathbf{X} = \mathbf{x} \cap \mathbf{Y} = \mathbf{y} \cap \mathbf{Z} = \mathbf{z}) = \mathrm{Pl}(\mathbf{Y}' = \mathbf{y}' | \mathbf{Y} = \mathbf{y} \cap \mathbf{Z} = \mathbf{z}).
$$

And since $I_{\mathrm{Pl}}^{rv}(\mathbf{X}, \mathbf{Y}|\mathbf{Z})$, it follows that $\mathrm{Pl}(\mathbf{Y} = \mathbf{y}|\mathbf{X} = \mathbf{x} \cap \mathbf{Z} = \mathbf{z}) = \mathrm{Pl}(\mathbf{Y} = \mathbf{y}|\mathbf{Z} = \mathbf{z})$. Plugging this into (10) and applying Alg2 again gives

$$
\begin{aligned}
&\mathrm{Pl}(\mathbf{Y} = \mathbf{y} \cap \mathbf{Y}' = \mathbf{y}'|\mathbf{X} = \mathbf{x} \cap \mathbf{Z} = \mathbf{z}) \\
={} &\mathrm{Pl}(\mathbf{Y}' = \mathbf{y}'|\mathbf{Y} = \mathbf{y} \cap \mathbf{Z} = \mathbf{z}) \otimes \mathrm{Pl}(\mathbf{Y} = \mathbf{y}|\mathbf{Z} = \mathbf{z}) \\
={} &\mathrm{Pl}(\mathbf{Y} = \mathbf{y} \cap \mathbf{Y}' = \mathbf{y}'|\mathbf{Z} = \mathbf{z}),
\end{aligned}
$$

as desired.

Now if $\mathrm{Pl}(\mathbf{Y} = \mathbf{y}|\mathbf{X} = \mathbf{x} \cap \mathbf{Z} = \mathbf{z}) = \bot$, then by CPl3, it follows that $\mathrm{Pl}(\mathbf{Y} = \mathbf{y} \cap \mathbf{Y}' = \mathbf{y}'|\mathbf{X} = \mathbf{x} \cap \mathbf{Z} = \mathbf{z}) = \bot$. Moreover, since $I_{\mathrm{Pl}}^{rv}(\mathbf{X}, \mathbf{Y}|\mathbf{Z})$, it follows that $\mathrm{Pl}(\mathbf{Y} = \mathbf{y}|\mathbf{Z} = \mathbf{z}) = \bot$. Applying CPl3, we get that $\mathrm{Pl}(\mathbf{Y} = \mathbf{y} \cap \mathbf{Y}' = \mathbf{y}'|\mathbf{Z} = \mathbf{z}) = \bot$. Thus, again 4(b) holds. ∎

**Theorem 5.6:** *If $(G, f)$ is $(D, \oplus, \otimes)$-representable then there is a unique cpm $\mathrm{Pl} \in \mathcal{PL}_{D, \oplus, \otimes}$ such that $(G, f)$ represents $\mathrm{Pl}$.*

**Proof:** Given $(G, f)$, suppose without loss of generality that $\mathbf{X} = \langle X_1, \ldots, X_n \rangle$ is a topological sort of the nodes in $G$. I now define the plausibility measure $\mathrm{Pl}$ determined by $(G, f)$. I start by defining $\mathrm{Pl}_{(G, f)}$ on sets of the form $\mathbf{X} = \mathbf{x}$.

It easily follows from Alg2 that if $\mathrm{Pl} \in \mathcal{PL}_{D, \oplus, \otimes}$ and $\mathrm{Pl}(X_1 = x_1 \cap \ldots \cap X_{n-1} = x_{n-1}) \neq \bot$, then

$$
\begin{aligned}
\mathrm{Pl}(\mathbf{X} = \mathbf{x}) ={} &\mathrm{Pl}(X_n = x_n|X_1 = x_1 \cap \ldots \cap X_{n-1} = x_{n-1}) \otimes \\
&\mathrm{Pl}(X_{n-1} = x_{n-1}|X_1 = x_1 \cap \ldots \cap X_{n-2} = x_{n-2}) \otimes \\
&\cdots \otimes \mathrm{Pl}(X_2 = x_2|X_1 = x_1) \otimes \mathrm{Pl}(X_1 = x_1).
\end{aligned}
\tag{11}
$$

Thus, an algebraic plausibility measure satisfies an analogue of the chain rule for probability. (Since $\otimes$ in $D$ is assumed to be associative, no parentheses are required here. However, even without this assumption, it follows easily from Alg2 that $\otimes$ is in fact associative on tuples $(a, b, c)$ of the form $(\mathrm{Pl}(U_1|U_2), \mathrm{Pl}(U_2|U_3), \mathrm{Pl}(U_3|U_4))$, where $U_1 \subseteq U_2 \subseteq U_3 \subseteq U_4$, which are the only types of tuples that arise in (11). Associativity will be more of an issue below.)

If $\mathrm{Pl}$ is compatible with $G$, then in fact

$$
\begin{aligned}
\mathrm{Pl}(\mathbf{X} = \mathbf{x}) ={} &\mathrm{Pl}(X_n = x_n|\cap_{X_j \in \mathrm{Par}_G(X_n)} X_j = x_j) \otimes \\
&\mathrm{Pl}(X_{n-1} = x_{n-1}|\cap_{X_j \in \mathrm{Par}_G(X_{n-1})} X_j = x_j) \otimes \\
&\cdots \otimes (X_1 = x_1).
\end{aligned}
\tag{12}
$$

(If $\mathrm{Par}_G(X_k) = \emptyset$, then $\mathrm{Pl}(X_k = x_k|\cap_{X_j \in \mathrm{Par}_G(X_k)} X_j = x_j)$ is just taken to be $\mathrm{Pl}(X_k = x_k)$.)

It is clear from (12) that $\mathrm{Pl}_{(G, f)}(\mathbf{X} = \mathbf{x})$ must be $d_{X_n, G, \mathbf{x}} \otimes \cdots \otimes d_{X_1, G, \mathbf{x}}$.

Note that every subset of $W$ can be written as a disjoint union of events of the form $\mathbf{X} = \mathbf{x}$. Thus, if $U \in \mathcal{F}$, define

$$
\mathrm{Pl}_{(G, f)}(U) = \oplus_{\{\mathbf{x}: \mathbf{X} = \mathbf{x} \subseteq U\}} d_{X_n, G, \mathbf{x}} \otimes \cdots \otimes d_{X_1, G, \mathbf{x}}.
$$

For conditional plausibilities, suppose that $\mathrm{Pl}_{(G, f)}(V) \neq \bot$, so that $V \in \mathcal{F}'$. Let $\{\mathbf{x}_1, \ldots, \mathbf{x}_k\} = \{\mathbf{x} : \mathbf{X} = \mathbf{x} \subseteq V\}$. It follows easily from BN6, BN7, R1, and R2 that $(\mathrm{Pl}_{(G, f)}(\mathbf{X} = \mathbf{x}_1), \ldots, \mathrm{Pl}_{(G, f)}(\mathbf{X} = \mathbf{x}_k)) \in D(\oplus)$. Thus, by BN8, if $\mathbf{X} = \mathbf{x} \subseteq V$, then $\mathrm{Pl}_{(G, f)}(\mathbf{X} = \mathbf{x}) \leq \mathrm{Pl}_{(G, f)}(V)$. By BN5, for each $j$, there exists $d_{\mathbf{X} = \mathbf{x}_j|V}$ such that

$$
(d_{\mathbf{X} = \mathbf{x}_j|V}, \mathrm{Pl}_{(G, f)}(V)) \in D(\otimes) \text{ and } d_{\mathbf{X} = \mathbf{x}_j|V} \otimes \mathrm{Pl}_{(G, f)}(V) = \mathrm{Pl}_{(G, f)}(\mathbf{X} = \mathbf{x});
$$

it follows from BN4 that $d_{\mathbf{X}=\mathbf{x}_j|V}$ is the unique element in $D$ with this property. Moreover, by BN5, $(d_{\mathbf{X}=\mathbf{x}_1|V}, \ldots, d_{\mathbf{X}=\mathbf{x}_k|V}) \in D(\oplus)$. Define $\mathrm{Pl}_{(G,f)}(U|V) = \oplus_{\{\mathbf{x}: \mathbf{X}=\mathbf{x} \subseteq U \cap V\}} d_{\mathbf{X}=\mathbf{x}|V}$ (where $\mathrm{Pl}_{(G,f)}(\emptyset|V)$ is taken to be $\perp$). Note for future reference that it follows from BN5 that $(\mathrm{Pl}_{(G,f)}(U|V), \mathrm{Pl}_{(G,f)}(V)) \in D(\otimes)$ and

$$\mathrm{Pl}_{(G,f)}(U|V) \otimes \mathrm{Pl}_{(G,f)}(V) = \mathrm{Pl}_{(G,f)}(U \cap V). \tag{13}$$

This completes the definition of $\mathrm{Pl}_{(G,f)}$. It remains to check that it is an algebraic cpm that is represented by $(G, f)$. Thus, we must check that Alg1–4 and CPl1–4 hold. Alg1 is immediate from the definitions and BN1 and BN2 (BN2 is necessary for the case that one of the disjoint sets is empty); Alg3 is immediate from BN3 and Alg4 is immediate from BN4. For Alg2, note that if $\mathrm{Pl}_{(G,f)}(V) \neq \perp$ and $\mathrm{Pl}_{(G,f)}(V') \neq \perp$ then, by (13), $\mathrm{Pl}_{(G,f)}(U \cap V|V') \otimes \mathrm{Pl}_{(G,f)}(V') = \mathrm{Pl}_{(G,f)}(U \cap V \cap V')$ and

$$
\begin{aligned}
&(\mathrm{Pl}_{(G,f)}(U|V \cap V') \otimes \mathrm{Pl}_{(G,f)}(V|V')) \otimes \mathrm{Pl}_{(G,f)}(V') \\
=\ & \mathrm{Pl}_{(G,f)}(U|V \cap V') \otimes (\mathrm{Pl}_{(G,f)}(V|V') \otimes \mathrm{Pl}_{(G,f)}(V')) \\
=\ & \mathrm{Pl}_{(G,f)}(U|V \cap V') \otimes \mathrm{Pl}_{(G,f)}(V \cap V') \\
=\ & \mathrm{Pl}_{(G,f)}(U \cap V \cap V').
\end{aligned}
$$

(Note that the associativity of $\otimes$ is being used here.) Thus, by BN4,

$$\mathrm{Pl}_{(G,f)}(U \cap V|V') = \mathrm{Pl}_{(G,f)}(U|V \cap V') \otimes \mathrm{Pl}_{(G,f)}(V|V').$$

CPl1 is immediate by definition (the empty sum is taken to be $\perp$). For CPl2, note that by (13), $\mathrm{Pl}_{(G,f)}(W|V) \otimes \mathrm{Pl}_{(G,f)}(V) = \mathrm{Pl}_{(G,f)}(V)$. Since $\top \otimes \mathrm{Pl}_{(G,f)}(V) = \mathrm{Pl}_{(G,f)}(V)$ by BN2, it follows from BN4 that $\mathrm{Pl}_{(G,f)}(W|V) = \top$. CPl3 follows readily from the definitions together with BN1, BN6, and BN7. CPl4 also follows by definition.

Next we must show that $(G, f)$ represents $\mathrm{Pl}_{(G,f)}$. The first step is to show that $\mathrm{Pl}_{(G,f)}(X = x|\mathrm{Par}_G(X) = \mathbf{z}) = d_{X=x|\mathrm{Par}_G(X)=\mathbf{z}}$. Note that by (13),

$$\mathrm{Pl}_{(G,f)}(X = x|\mathrm{Par}_G(X) = \mathbf{z}) \otimes \mathrm{Pl}_{(G,f)}(\mathrm{Par}_G(X) = \mathbf{z}) = \mathrm{Pl}_{(G,f)}(X = x \cap \mathrm{Par}_G(X) = \mathbf{z}).$$

By definition,

$$\mathrm{Pl}_{(G,f)}(X = x \cap \mathrm{Par}_G(X) = \mathbf{z}) = \oplus_{\{\mathbf{x}: \mathbf{X}=\mathbf{x}' \subseteq (X=x \cap \mathrm{Par}_G(X)=\vec{y})\}} \mathrm{Pl}_{(G,f)}(\mathbf{X} = \mathbf{x}').$$

Each term in the "sum" on the right is the "product" of terms; indeed, the sum is over all possible products that include $d_{X=y|\mathrm{Par}_G(X)=\mathbf{z}}$ as one of the terms and a term $d_{Y=y|\mathrm{Par}_G(Y)=\mathbf{z}'}$ for each $Y \in \mathrm{Par}_G(X)$, where $y$ is the component of $\mathbf{z}$ corresponding to $Y$. By using BN1, BN3, R1, and R2, it is not hard to show that

$$
\begin{aligned}
& \mathrm{Pl}_{(G,f)}(X = y \cap \mathrm{Par}_G(X) = \mathbf{z}) \\
=\ & \oplus_{\{: \mathbf{X}=\mathbf{x}' \subseteq (X=x \cap \mathrm{Par}_G(X)=\vec{y})\}} \mathrm{Pl}_{(G,f)}(\mathbf{X} = \mathbf{x}') \\
=\ & d_{X=x|\mathrm{Par}_G=\mathbf{z}} \otimes \mathrm{Pl}_{(G,f)}(\mathrm{Par}_G(X) = \mathbf{z}).
\end{aligned}
\tag{14}
$$

It now follows from BN4 that $\mathrm{Pl}_{(G,f)}(X = x|\mathrm{Par}_G(X) = \mathbf{z}) = d_{X=x|\mathrm{Par}_G(X)=\mathbf{z}}$.

To show that $\mathrm{Pl}_{(G,f)}(X = x|\mathrm{ND}_G(X) = \vec{y} \cap \mathrm{Par}_G(X) = \mathbf{z}) = d_{X=x|\mathrm{Par}_G(X)=\mathbf{z}}$, it suffices to show that

$$
\begin{aligned}
& \mathrm{Pl}_{(G,f)}(X = x \cap \mathrm{ND}_G(X) = \vec{y} \cap \mathrm{Par}_G(X) = \mathbf{z}) \\
=\ & d_{X=x|\mathrm{Par}_G(X)=\mathbf{z}} \otimes \mathrm{Pl}_{(G,f)}(\mathrm{ND}_G(X) = \vec{y} \cap \mathrm{Par}_G(X) = \mathbf{z}),
\end{aligned}
\tag{15}
$$

for then the result follows by BN5. (15) can be shown much like (14), but now the commutativity of $\otimes$ (BN1) is essential. That is, the expressions for $\mathrm{Pl}_{(G,f)}(X = x \cap \mathrm{ND}_G(X) = \vec{y} \cap \mathrm{Par}_G(X) = \mathbf{z})$ and $d_{X=x|\mathrm{Par}_G(X)=\mathbf{z}} \otimes \mathrm{Pl}_{(G,f)}(\mathrm{ND}_G(X) = \vec{y} \cap \mathrm{Par}_G(X) = \mathbf{z})$ involve the same terms, but not necessarily in the same order. With commutativity, they can be permuted so that they are in the same order.

Similar arguments, which I leave to the reader, show that $\mathrm{Pl}_{(G,f)}(\mathrm{ND}_G(X) = \vec{y}|X = x \cap \mathrm{Par}_G(X) = \mathbf{z}) = \mathrm{Pl}_{(G,f)}(\mathrm{ND}_G(X) = \vec{y}|\mathrm{Par}_G(X) = \mathbf{z})$. Thus, $(G, f)$ represents $\mathrm{Pl}_{(G,f)}$. ∎

**Theorem 5.10:** *Suppose that $(D, \oplus, \otimes)$ is a rich BN-compatible domain. Then if $I_{\mathrm{Pl}}^{rv}(\mathbf{X}, \mathbf{Y}|\mathbf{Z})$ holds for every plausibility measure in $\mathcal{PL}_{D,\oplus,\otimes}$ compatible with $G$, then $d\text{-}sep_G(\mathbf{X}, \mathbf{Y}|\mathbf{Z})$.*

**Proof:** Suppose that $\mathbf{X}$ is not d-separated from $\mathbf{Y}$ by $\mathbf{Z}$ in $G$. Then there is some $X \in \mathbf{X}$ and $Y \in \mathbf{Y}$ such that $X$ is not d-separated from Y by $\mathbf{Z}$ in $G$. I construct a cpm in $\mathrm{Pl} \in \mathcal{PL}_{D,\oplus,\otimes}$ such that $I_{\mathrm{Pl}}^{rv}(X, Y|\mathbf{Z})$ does not hold, using the techniques of (Geiger et al., 1990).

As shown in (Geiger et al., 1990, Lemma 9), if $X$ is not d-separated from $Y$ in $G$, there exists a subgraph $G'$ of $G$ such that

1. $G'$ includes all the nodes in $G$ but only a subset of the edges in $G$,

2. $X$ is not d-separated from $Y$ by $Z$ in $G'$.

3. the edges $E'$ in $G'$ consist only of those specified below:

   (a) a *trail* $q$ from $X$ to $Y$,
   (b) for every head-to-head node $X_i$ on the trail $q$, there is a directed path $p_i$ in $G'$ to a node in $\mathbf{Z}$; moreover, the paths $p_i$ do not share any nodes and the only node that $p_i$ shares with $q$ is $X_i$.

Note that every node in $G'$ has either 0, 1, or 2 parents in $G'$. Let $(G', f)$ be a quantitative Bayesian network such that for each node in $X$ in $G'$ with no parents in $G'$, the cpt $f(X)$ is such that $d_{X=0} = d$ and $d_{X=1} = d'$. If a node $X$ in $G'$ has one parent, say $X'$, then the cpt $f(X)$ is such that $d_{X=i|X'=j}$ is $\top$ if $i = j$ and $\perp$ if $i \neq j$. Finally, if $X$ has two parents, say $X'$ and $X''$, the cpt $f(X)$ is such that $d_{X=k|X'=i\cap X''=k}$ is $\top$ if $k = i \oplus j \pmod 2$ and is $\perp$ otherwise. Since $d \oplus d' = \top$ and BN2 guarantees that $\top \oplus \perp = \top$, the construction satisfies R1. The richness of $D$ guarantees that R2 holds. By Theorem 5.6, there is a (unique) plausibility measure in $\mathrm{Pl} \in \mathcal{PL}_{D,\oplus,\otimes}$ that is represented by $(G', f)$. It is easy to check that Pl is compatible with $G$ as well. There are three cases to consider:

- Suppose that $X$ has no parents in $G'$. Then it is easy to see that $I_{\mathrm{Pl}}^{rv}(X, \mathbf{Y}|\mathbf{Z})$ for all $\mathbf{Y}$ and $\mathbf{Z}$ (and, in particular, if $\mathbf{Y} = \mathrm{ND}_G(X)$ and $\mathbf{Z} = \mathrm{Par}_G(X)$).

- Suppose that $X$ has one parent in $G'$, say $X'$. Then it is easy to see that $I_{\mathrm{Pl}}^{rv}(X, \mathbf{Y}|\mathbf{Z})$ holds for all $\mathbf{Y}$ and $\mathbf{Z}$ such that $X' \in \mathbf{Z}$. Since $X'$ is a parent of $X$ in $G$, again $I_{\mathrm{Pl}}^{rv}(X, \mathrm{ND}_G(X)|\mathrm{Par}_G(X))$ must hold.

- Finally, if $X$ has two parents in $G'$, say $X'$ and $X''$, then it is easy to see that $I_{\mathrm{Pl}}^{rv}(X, \mathbf{Y}|\mathbf{Z})$ holds for all $\mathbf{Y}$ and $\mathbf{Z}$ such that $\{X', X''\} \subseteq \mathbf{Z}$. Since $X'$ and $X''$ are parents of $X$ in $G$, again $I_{\mathrm{Pl}}^{rv}(X, \mathrm{ND}_G(X)|\mathrm{Par}_G(X))$ must hold. ∎

## Acknowledgments

A preliminary version of this paper appears in *Uncertainty in Artificial Intelligence, Proceedings of the Sixteenth Conference*, 2000. I thank Serafín Moral, Fabio Cozman, Peter Walley, and the anonymous referees of both the UAI and journal version of the paper for very useful comments. This work was supported in part by the NSF, under grants IRI-96-25901 and IIS-0090145.

## References

Campos, L., & Huete, J. F. (1993). Independence concepts in upper and lower probabilities. In Bouchon-Meunier, B., Valverde, L., & Yager, R. R. (Eds.), *Uncertainty in Intelligent Systems*, pp. 85–96. North-Holland, Amsterdam.

Campos, L., & Huete, J. F. (1999a). Independence concepts in possibility theory: Part I. *Fuzzy Sets and Systems*, *103*(1), 127–152.

Campos, L., & Huete, J. F. (1999b). Independence concepts in possibility theory: Part II. *Fuzzy Sets and Systems*, *103*(3), 487–505.

Campos, L., & Moral, S. (1995). Independence concepts for sets of probabilities. In *Proc. Eleventh Conference on Uncertainty in Artificial Intelligence (UAI '95)*, pp. 108–115.

Cousa, I., Moral, S., & Walley, P. (1999). Examples of independence for imprecise probabilities. In *Proc. First Intl. Symp. Imprecise Probabilities and Their Applications*.

Cozman, F. G. (1998). Irrelevance and independence relations in Quasi-Bayesian networks. In *Proc. Fourteenth Conference on Uncertainty in Artificial Intelligence (UAI '98)*, pp. 89–96.

Cozman, F. G. (2000a). Credal networks. *Artificial Intelligence*, *120*(2), 199–233.

Cozman, F. G. (2000b). Separation properties of setes of probability measures. In *Proc. Sixteenth Conference on Uncertainty in Artificial Intelligence (UAI 2000)*.

Cozman, F. G., & Walley, P. (1999). Graphoid properties of epistemic irrelevance and independence. Unpublished manuscript.

Darwiche, A. (1992). *A Symbolic Generalization of Probability Theory*. Ph.D. thesis, Stanford University.

Darwiche, A., & Ginsberg, M. L. (1992). A symbolic generalization of probability theory. In *Proceedings, Tenth National Conference on Artificial Intelligence (AAAI '92)*, pp. 622–627.

Darwiche, A., & Goldszmidt, M. (1994). On the relation between kappa calculus and probabilistic reasoning. In *Proc. Tenth Conference on Uncertainty in Artificial Intelligence (UAI '94)*, pp. 145–153.

Dubois, D., Fariñas del Cerro, L., Herzig, A., & Prade, H. (1994). An ordinal view of independence with applications to plausible reasoning. In *Proc. Tenth Conference on Uncertainty in Artificial Intelligence (UAI '94)*, pp. 195–203.

Dubois, D., & Prade, H. (1990). An introduction to possibilistic and fuzzy logics. In Shafer, G., & Pearl, J. (Eds.), *Readings in Uncertain Reasoning*, pp. 742–761. Morgan Kaufmann, San Francisco, Calif.

Fagin, R., & Halpern, J. Y. (1991). A new approach to updating beliefs. In Bonissone, P., Henrion, M., Kanal, L., & Lemmer, J. (Eds.), *Uncertainty in Artificial Intelligence 6*, pp. 347–374. Elsevier Science Publishers, Amsterdam.

Finetti, B. d. (1936). Les probabilités nulles. *Bulletins des Science Mathématiques (première partie)*, *60*, 275–288.

Fonck, P. (1994). Conditional independence in possibility theory. In *Proc. Tenth Conference on Uncertainty in Artificial Intelligence (UAI '94)*, pp. 221–226.

Friedman, N., & Halpern, J. Y. (1995). Plausibility measures: a user's guide. In *Proc. Eleventh Conference on Uncertainty in Artificial Intelligence (UAI '95)*, pp. 175–184.

Geiger, D., & Pearl, J. (1988). On the logic of causal models. In *Proc. Fourth Workshop on Uncertainty in Artificial Intelligence (UAI '88)*, pp. 136–147.

Geiger, D., Verma, T., & Pearl, J. (1990). Identifying independence in bayesian networks. *Networks*, *20*, 507–534.

Gilboa, I., & Schmeidler, D. (1993). Updating ambiguous beliefs. *Journal of Economic Theory*, *59*, 33–49.

Goldszmidt, M., & Pearl, J. (1992). Rank-based systems: A simple approach to belief revision, belief update and reasoning about evidence and actions. In *Principles of Knowledge Representation and Reasoning: Proc. Third International Conference (KR '92)*, pp. 661–672. Morgan Kaufmann, San Francisco, Calif.

Halpern, J. Y. (2000). Conditional plausibility measures and Bayesian networks. In *Proc. Sixteenth Conference on Uncertainty in Artificial Intelligence (UAI 2000)*, pp. 247–255. To appear, *Journal of A.I. Research*.

Keynes, J. M. (1921). *A Treatise on Probability*. Macmillan, London.

Levi, I. (1985). Imprecision and uncertainty in probability judgment. *Philosophy of Science*, *52*, 390–406.

Pearl, J. (1988). *Probabilistic Reasoning in Intelligent Systems*. Morgan Kaufmann, San Francisco, Calif.

Popper, K. R. (1968). *The Logic of Scientific Discovery (revised edition)*. Hutchison, London. The first version of this book appeared as *Logik der Forschung*, 1934.

Rényi, A. (1964). Sur les espaces simples de probabilités conditionelles. *Annales de l'Institut Henri Poincaré, Nouvelle série, Section B, 1*, 3–21. Reprinted as paper 237 in *Selected Papers of Alfred Rényi, III: 1962–1970*, Akadémia Kiadó, 1976, pp. 284–302.

Shafer, G. (1976). *A Mathematical Theory of Evidence*. Princeton University Press, Princeton, N.J.

Shenoy, P. P. (1994). Conditional independence in valuation based systems. *International Journal of Approximate Reasoning, 10*, 203–234.

Shenoy, P. P., & Shafer, G. (1990). An axiomatic framework for Bayesian and belief-function propagation. In Shachter, R., Levitt, T., Kanal, L., & Lemmer, J. (Eds.), *Uncertainty in Artificial Intelligence 4*, pp. 169–198.

Spohn, W. (1988). Ordinal conditional functions: a dynamic theory of epistemic states. In Harper, W., & Skyrms, B. (Eds.), *Causation in Decision, Belief Change, and Statistics*, Vol. 2, pp. 105–134. Reidel, Dordrecht, Netherlands.

Verma, T. (1986). Causal networks: semantics and expressiveness. Technical report R–103, UCLA Cognitive Systems Laboratory.

Walley, P. (1991). *Statistical Reasoning with Imprecise Probabilities*, Vol. 42 of *Monographs on Statistics and Applied Probability*. Chapman and Hall, London.

Wang, Z., & Klir, G. J. (1992). *Fuzzy Measure Theory*. Plenum Press, New York.

Weydert, E. (1994). General belief measures. In *Proc. Tenth Conference on Uncertainty in Artificial Intelligence (UAI '94)*, pp. 575–582.

Wilson, N. (1994). Generating graphoids from generalized conditional probability. In *Proc. Tenth Conference on Uncertainty in Artificial Intelligence (UAI '94)*, pp. 583–591.

Zadeh, L. A. (1978). Fuzzy sets as a basis for a theory of possibility. *Fuzzy Sets and Systems, 1*, 3–28.